\documentclass{article}

\usepackage{arxiv}

\usepackage[utf8]{inputenc} % allow utf-8 input
\usepackage[T1]{fontenc}    % use 8-bit T1 fonts
\usepackage{hyperref}       % hyperlinks
\usepackage{url}            % simple URL typesetting
\usepackage{booktabs}       % professional-quality tables
\usepackage{amsfonts}       % blackboard math symbols
\usepackage{nicefrac}       % compact symbols for 1/2, etc.
\usepackage{microtype}      % microtypography
\usepackage{lipsum}		% Can be removed after putting your text content

\usepackage{graphicx}
\usepackage{floatrow}
\usepackage{doi}

\usepackage{wrapfig}
\usepackage{multirow}
\usepackage{algorithmicx}
\usepackage{algpseudocode}
\usepackage{algcompatible}
\usepackage[ruled]{algorithm}
\usepackage{amsmath}
\usepackage{xspace}
\usepackage{subcaption}
\usepackage[normalem]{ulem}
\usepackage{bbm}
\usepackage{tikz}
\usetikzlibrary{decorations.fractals,spy}

\title{Taming 3DGS: High-Quality Radiance Fields with Limited Resources}
\renewcommand{\undertitle}{}
\renewcommand{\headeright}{}

\date{} 					% Or removing it

\author{%
  \begin{tabular}[t]{c}
    Saswat Subhajyoti Mallick\thanks{Equal contribution} $^{\, 1}$ \quad
    Rahul Goel$^{*\, 3}$  \quad
    Bernhard Kerbl$^1$   \\
    Francisco Vicente Carrasco$^1$ \quad 
    Markus Steinberger$^2$ \quad
    Fernando De La Torre$^1$ \\
    \end{tabular}%
  \quad
  \and
  \begin{tabular}[t]{c}
    $^1$Carnegie Mellon University\quad 
    $^2$Graz University of Technology\\
    $^3$International Institute of Information Technology, Hyderabad\\
  \end{tabular}%
}

\begin{document}
\maketitle

\begin{abstract}
3D Gaussian Splatting (3DGS) has transformed novel-view synthesis with its fast, interpretable, and high-fidelity rendering. However, its resource requirements limit its usability. Especially on constrained devices, training performance degrades quickly and often cannot complete due to excessive memory consumption of the model. The method converges with an indefinite number of Gaussians---many of them redundant---making rendering unnecessarily slow and preventing its usage in downstream tasks that expect fixed-size inputs.

To address these issues, we tackle the challenges of training and rendering 3DGS models on a budget. We use a guided, purely constructive densification process that steers densification toward Gaussians that raise the reconstruction quality. Model size continuously increases in a controlled manner towards an exact budget, using score-based densification of Gaussians with training-time priors that measure their contribution.

We further address training speed obstacles: following a careful analysis of 3DGS' original pipeline, we derive faster, numerically equivalent solutions for gradient computation and attribute updates, including an alternative parallelization for efficient backpropagation. We also propose quality-preserving approximations where suitable to reduce training time even further. Taken together, these enhancements yield a robust, scalable solution with reduced training times, lower compute and memory requirements, and high quality. Our evaluation shows that in a budgeted setting, we obtain competitive quality metrics with 3DGS while achieving a 4--5$\times$ reduction in both model size and training time. With more generous budgets, our measured quality surpasses theirs. These advances open the door for novel-view synthesis in constrained environments, e.g., mobile devices. 

\end{abstract}


\section{Introduction}
\label{sec:intro}
%Novel view synthesis (NVS) tackles the challenge of predicting unseen views from a set of calibrated, multi-view images. 
Novel View Synthesis (NVS) predicts unseen views from multi-view datasets, enabling users to freely explore 3D content from as little as a handful of easy-to-obtain photographs.
State-of-the-art NVS solutions can yield photo-realistic results that produce high-quality user experiences for e-commerce, entertainment, and immersive telecommunication. Recently, NVS methods have also emerged as a powerful conditioning tool for high-quality 3D surface reconstruction. 
The extensive research body on NVS covers various methodologies, ranging from image- and mesh-based to purely neural representations. Within this domain, 3D Gaussian Splatting (3DGS) has been gaining popularity, since it combines high-quality image synthesis, fast real-time rendering, and amenable training times~\cite{kerbl20233d}.
%prominent technique.  ********* We need to talk a bit more about application of NVS  and say something about Nerf limitation***********
%****** This paragraph is all about limitations of existing techniques******
%To judge the general usability of an NVS method, we must consider its overall resource requirements for both training and inference. 
3DGS leverages an explicit, point-based scene representation, a differentiable rendering pipeline, and GPU-optimized rasterization to achieve photo-realistic image synthesis at high frame rates. However, its optimization procedure is difficult to control; this process---although it includes several heuristics---is often wasteful and can lead to excessive memory overheads.

Starting from a sparse set of input points, many of the eventual optimized primitives are redundant: Gaussians often make only minor contributions in areas where fewer would suffice, while other regions remain under-reconstructed and blurry. This inefficient distribution of Gaussian primitives impacts not only training time but also the practical aspects of the representation. A typical 3DGS model can yield several millions of Gaussians for a single unbounded scene and require more than one gigabyte of disk space. 
Such substantial memory usage and geometry workload complicate real-time rendering on low-end devices, preventing application in constrained settings like network streaming or AR/VR on embedded systems. 

In addition to being excessive, the memory consumption of 3DGS is also hard to predict: even when starting from the same number of input points, the difference between two reconstructed scenes w.r.t.\ the number of Gaussians (and thus required storage) can be as much as one order of magnitude. 
This hinders its usability for downstream applications that require a fixed number of inputs (e.g., classifier networks), preventing them from exploiting an otherwise efficient and explicit representation. 
Similarly, training time---although acceptable---fluctuates strongly and overall fails to reflect the much higher rendering speed of 3DGS.
%Unfortunately, this makes it impossible to predict whether 3DGS training will succeed with any given hardware or time budget. 

\begin{figure}
    \includegraphics[width=0.69\textwidth]{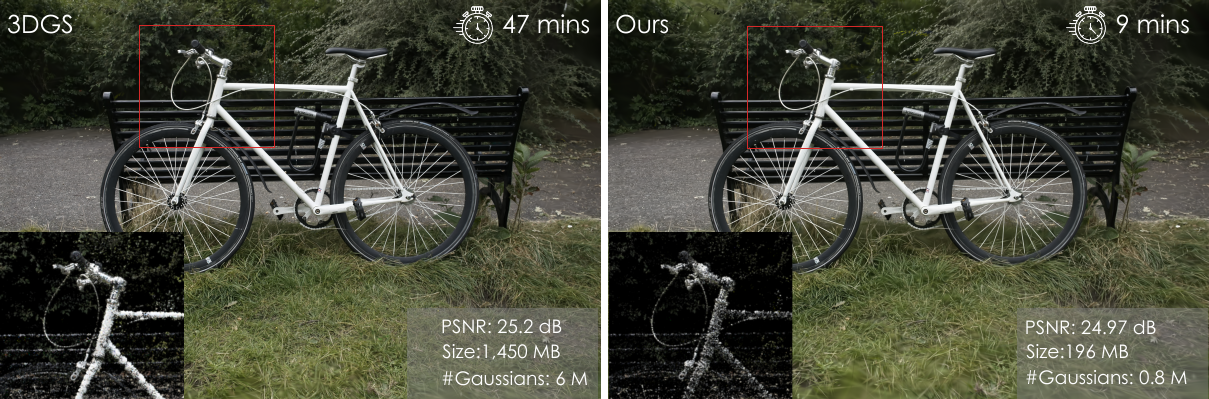}
    \hfill
    \includegraphics[width=0.29\textwidth]{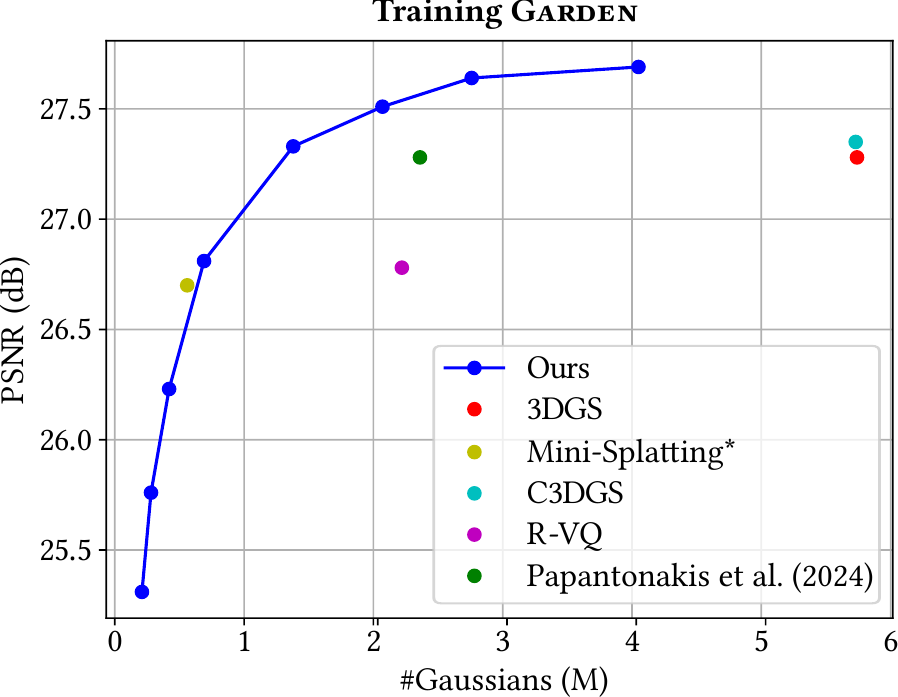}
    \caption{Our method makes 3DGS optimization fast and flexible, achieving high rendering quality on a budget. Left and middle: model size and training time are reduced by more than $5\times$. Right: Our method produces models with an exact, user-specified target size, surpassing 3DGS quality as the target increases.}
    \label{fig:teaser}
\end{figure}

In order to tame 3DGS, we propose a strict moderation in the Gaussian densification process to provide close control over its resource consumption (see Fig.~\ref{fig:teaser}). Given a user-defined model size, we ensure a deterministic training schedule that can yield the exact number of desired Gaussians. To achieve high quality with fewer primitives (less than 5$\times$ on average), we tackle the suboptimal distribution and high redundancy of the original method. We propose an alternate densification algorithm, guided by a flexible, score-based sampling of Gaussian primitives. Our suggested scoring scheme for high quality at a budget combines loss-relevant components that we collect per Gaussian, and across multiple sampled training views. Densification occurs according to the pre-defined budget in the vicinity of the top-scoring Gaussians. In contrast to previous work, our densification uses a \emph{purely constructive} schedule: we do not require substantial pruning or culling of Gaussians during training. Therefore, we avoid unnecessary peaks in the optimization that could violate the user's hardware or budget constraints. %\fvnote{Thus, facilitating training in embeded devices.}
%A side benefit of limiting model size is increased training speed. However, 

Redundancy in 3DGS is not limited to its eventual primitive distribution. Therefore, we closely analyze the time cost and quality tradeoff for individual steps in the training pipeline and propose alternative, more efficient substitutes. This includes revisiting the parallelization opportunities of backpropagation, which we change from a per-pixel to a per-splat approach. We achieve a significant speedup of 4--5$\times$ on average (see Fig.~\ref{fig:teaser}), with training times reduced to just a few minutes on consumer-grade hardware. %The notion of \textit{budget} signifies how detailed does one need the scene to be. 
%\ssnote{Add some numbers as to how much budget is usually enough, how much reduction, improvement in fps we bring, etc.} 
%To the best of our knowledge, we're the first to propose an alternate densification approach with rigid limits on model size, making 3DGS hardware-friendly, without any additional post-processing. \\
Our contributions to taming 3DGS can thus be summarized as follows:
\begin{enumerate}
	\item \emph{A purely constructive, budget-constrained optimization}, enabling full control over model size and resources.
	\item \emph{A flexible framework for score-based densification}, allowing for use case-specific behavior and prioritization, e.g., by indicating important regions of interest.
	\item \emph{Analysis and significant speedup of relevant training steps}, using both equivalent and approximate substitute methods.
\end{enumerate}

\section{Related work}
An extensive body of previous work focuses on novel-view synthesis: we first provide a brief overview of the most common approaches to this problem, before delving into solutions that focus specifically on raising the efficiency and portability of 3D Gaussian Splatting. Finally, we discuss point cloud downsampling approaches, from which we draw inspiration in our score-based 
densification.

\textbf{Novel-View Synthesis.}
Previous work has explored a wide range of solutions for reconstructing or predicting the appearance of scenes, ranging from small-scale models~\cite{chaurasia2013depth,buehler2023unstructured,Nishant2023SVS2} to unbounded environments~\cite{Riegler2021SVS,DeepBlending2018,bodis2016efficient}.
In contrast, Neural Radiance Fields (NeRFs)~\cite{mildenhall2021nerf} use an implicit representation, which is trained using gradient descent to recover a volumetric, continuous radiance field. While the initially proposed method was limited to single objects---taking over a day to process them---several follow-up works raised the scope and speed of NeRF scene reconstruction~\cite{barron2021mip,kaizhang2020nerfpp,barron2022mipnerf360,tensorf}. To address the high rendering times, voxel-based representations~\cite{Sun_2022_DVGO, karnewar2022relu} have been proposed to complement or replace select components of the NeRF architecture. 

Significant breakthroughs for both training and rendering performance were marked by the use of hash grids~\cite{muller2022instant} and space warping~\cite{wang2023f2nerf}, at the cost of introducing quality caps.
State-of-the-art NeRF-based techniques ~\cite{barron2023zipnerf, duckworth2023smerf, wu2022snisr, zhang2022nerfusion, niemeyer2024radsplat} are capable of reconstructing unbounded scenes with high quality and render at interactive frame rates, however, training them requires significant time and compute effort. The recently introduced 3D Gaussian Splatting (3DGS) uses an initial point cloud---a common side product of calibration---and converts it to optimizable 3D Gaussian primitives~\cite{kerbl20233d}. 3DGS achieves high quality and extremely fast rendering; however, it suffers from exorbitant, unpredictable storage demands and fluctuating training times, making it a poor choice for performing novel-view synthesis \emph{at a budget}.

\begin{figure*}
    \begin{subfigure}{0.81\textwidth}
    \includegraphics[height=4.8cm]{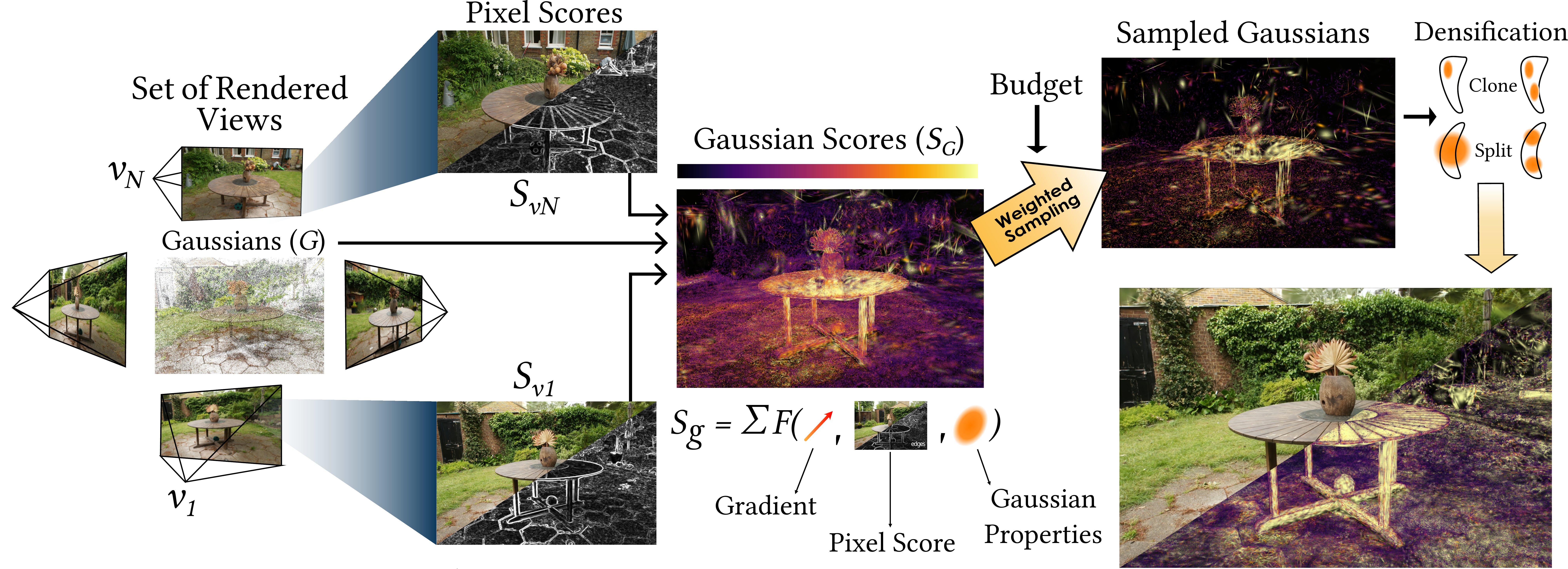}
    \caption{Score-based Sampling for Densification}
    \label{fig:overview}
    \end{subfigure}
    \hfill
    \begin{subfigure}{0.18\textwidth}
    \includegraphics[height=4.8cm]{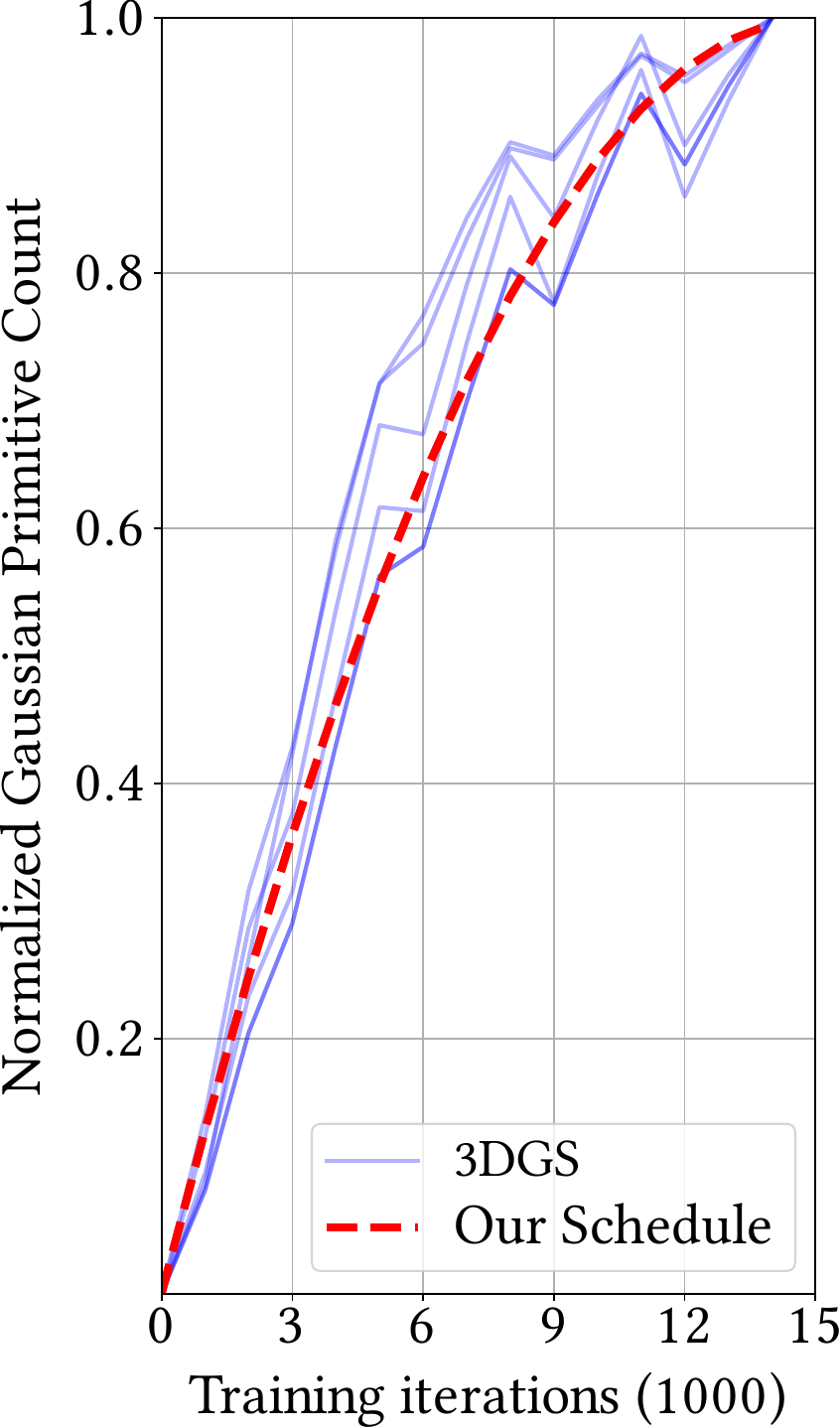}
    \caption{Predictable Growth}
    \label{fig:curve}
    \end{subfigure}
    \caption{Overview of our method. (a) We propose a systematic redesign of 3DGS densification. To select Gaussians to densify, we sample training views and compute per-pixel saliency. A scoring function $F$ combines gradient, saliency, and primitive properties into a per-Gaussian score $\textbf{S}_g$. (b) The addition of new Gaussians follows a predictable schedule. We follow a growth curve that mimics 3DGS' behavior and can be fitted to yield any desired model size after training.}
\end{figure*}

\textbf{3D Gaussian Splatting and Compression.} Several recent works have managed to considerably reduce the on-disk storage requirements of 3DGS. Compressing a model's feature space is a widely adopted technique~\cite{navaneet2023compact3d, lee2023compact}; the parameters of the Gaussians (geometry, color, opacity) can be clustered and indexed using codebooks. This reduces the compute and storage footprint per primitive, alleviating total memory consumption without significant quality degradation. Niedermayr et. al.~\cite{niedermayr2023compressed} follow a similar recipe, but use thorough, sensitivity-aware clustering on Gaussian parameters, followed by a quantization-aware fine-tuning and entropy encoding. Fan et. al.~\cite{fan2023lightgaussian} weight Gaussians on their volume and opacity to prune the less significant ones, followed by distillation from synthetic (pseudo-)views and quantization of parameters. 
Papantonakis et. al.~\cite{papanto2024} cull Gaussian primitives based on their spatial density and adaptively prune view-dependent color coefficients on demand.

While these methods are effective in reducing the storage requirements of 3DGS, they do little to make the process more 
\emph{controllable}. Furthermore, although several approaches consider the decimation of Gaussian primitives, they usually cause modest reductions of ${\approx}2\times$.
Other aspects of previously proposed on-disk compression techniques, such as code-booking or entropy minimization, are directly compatible with our method, which would lead to even smaller file sizes due to our higher primitive reduction.

\textbf{Point Cloud Downsampling.}
\label{subsec:pcd_sampling} 
By interpreting Gaussian means as singular points in space, we find that optimizing for high quality at low primitive counts is closely related to \emph{point cloud downsampling}. 
Point clouds are 3D points distributed in space, often representing surfaces or the density of measured objects. %They are used for developing 3D assests in multitude of animating, rendering, and mass customization applications. 
Especially when resulting from real-world scanning, the considerable size of point cloud data can become a computation burden. This causes setbacks for downstream applications running on compute-constrained hardware settings. 
Previous work addresses this problem by quantizing the space and approximating samples using nearest neighbors~\cite{goldberger2004neighbourhood, plotz2018neural}, resampling points based on their density and distribution. 

Learning-based methods introduce task-specific sampling~\cite{dovrat2019learning} and yield results competitive with heuristic methods, such as farthest point sampling. Nezhadarya et. al.~\cite{nezhadarya2020adaptive} uses a critical points layer, which qualifies the most significant points to the next network layer. Yang et. al.~\cite{yang2019modeling} implement Gumbel subset sampling to improve the classification accuracy of a network trained on point cloud data. Lang et. al.~\cite{lang2020samplenet} introduce a differentiable projection during nearest-neighbor search that "softens" the discrete points. Inspired by these sampling-based approaches to produce compact, yet salient datasets, we redesign 3DGS densification as a sampling-guided procedure.

\section{Method}
Our approach is outlined in Fig.~\ref{fig:overview}. SfM point clouds are used as an initialization to train a 3DGS-based model from calibrated multi-view images with a pre-determined densification schedule. The original 3DGS densification algorithm continuously adds primitives (details) to regions with high positional gradients, splitting large Gaussians, cloning smaller ones, and removing transparent ones. We replace this module with a less frequently executed procedure built upon steerable sampling. The maximum number of new Gaussians added at every stage is pre-determined: although our method mimics the original 3DGS growth curve, the peak (and final) number of Gaussians is fully controllable by the user who provides the limits for model size (Fig.~\ref{fig:curve}). To maximize the quality per Gaussian, our densification is guided using a score-based ranking and employs \emph{high-opacity Gaussians} to increase the primitives' expressiveness. In addition, training duration is significantly reduced through several proposed modifications that target the primary bottlenecks of the original pipeline, including a faster, numerically equivalent solution for backpropagation. Taken together, these measures yield an optimization with high controllability, flexibility, and performance. 
%Our solution allows for custom, specialized formulations of importance; our \emph{suggested} importance formulation reveals that we can use only 15\% of the Gaussians of the original 3DGS to reconstruct a scene without any noticeable degradation in quality.

\subsection{3D Gaussian Splatting Background}
3DGS~\cite{kerbl20233d} is a point-based approach that models scenes using a set of 3D Gaussians, parameterized by position ($\mu$), covariance ($\Sigma$), and opacity $o$. Ignoring inter-primitive overlap, the theoretical contribution of a 3D Gaussian at a point $x$ is defined by:

\begin{equation}
	G(x) = oe^{-\frac{1}{2}(x-\mu)^T\Sigma^{-1}(x-\mu)}, \quad \Sigma=RSS^TR^T,
\end{equation}

where $R$ is a rotation and $S$ a scaling matrix.
View-dependent appearance is modeled by Spherical Harmonics (SH) of order $3$ and a direct color component for base appearance. For a particular viewpoint, the visible set of 3D Gaussians  %that fall within the view frustum 
is rendered in a tile-based, differentiable rasterizer to obtain a 2D image by $\alpha$-blending their projections (splats). 
3DGS training minimizes a combined \textit{$L_1$} and SSIM loss w.r.t.\ the rendered and ground truth image by optimizing the parameters---position, rotation, scaling, opacity, and SH---of each Gaussian. 

\subsection{Predictable Model Growth}
\label{subsec:growth_curve}

Throughout optimization, 3DGS continuously \emph{densifies} its representation by adding Gaussian primitives to resolve under-reconstructed regions. 
However, the number of added primitives at each stage is decided based on a simple thresholding operation, with no control over the progressive or final count. This evolutionary automaton---although effective---leads to hard-to-predict, often exorbitant model sizes and fluctuating training times.

To define a simpler, yet similarly effective and fully predictable growth pattern, we investigate the densification behavior of 3DGS across the outdoor scenes in the MipNeRF360 dataset. Fig.~\ref{fig:curve} plots the development in the number of total Gaussians for each scene as training progresses with the original method; note that curves have been renormalized on the range between their initial and final 3DGS primitive count. 
We find that the number of Gaussians added in each step follows a trend of quadratic decrease. 
%This is because the positional lr decays in a similar fashion and hence, fewer gaussians qualify for getting densified over the iterations.
We exploit this pattern to determine a schedule of added primitives at each step, using a parabolic curve that starts from the SfM initialization and peaks precisely at the \emph{user-defined budget}: %The exact growth function $A$ is outlined below:

\begin{equation}
	A(x) = \frac{B-S-2N}{N^2}x^2 + 2x + B,
\end{equation}

where $N$ is the number of densification steps, $B$ is the final count (budget), and $S$ is the number of SfM points at initialization. Since 3DGS prunes low-opacity Gaussians over time, following an additive schedule directly may produce fewer primitives than the given target. To avoid this, we instead compute the difference between our current and \emph{accumulated} target count and densify the corresponding number of primitives. Sec.~\ref{sec:eval} demonstrates the effectiveness of this scheme and the graceful quality degradation resulting from lower budget limits.

\subsection{Steerable Densification with Sampling}
The original 3DGS approach suggests that high positional gradients on a Gaussian indicate insufficient samples in its vicinity. Hence, such Gaussians are regularly densified, either by \emph{cloning} or \emph{splitting} (depending on their size).
Bleeding-edge research reformulated the 3DGS optimization process as a sequence of Stochastic Langevin Gradient Descent (SLGD) updates \cite{kheradmand20243d}.
At any point, the optimized set of Gaussians can be interpreted as samples from a likelihood distribution tied to 3DGS' overall loss.
Obtaining a complete, high-fidelity reconstruction demands a solution that delicately balances optimization and exploration.
Letting image loss also steer the densification procedure seems intuitive: a high loss can indicate the need for denser sampling or additional exploration.

In the spirit of maintaining a steerable, yet interpretable densification procedure, we propose a flexible solution that incorporates salient indicators like image loss directly into the process. 
This is enabled via two key features: a score-based, customizable sampling of densification candidates and a significantly reduced densification frequency.
The former combines salient per-Gaussian and per-pixel metrics, such as loss, to decide each primitive's probability of densification. 

The reduction in densification frequency is motivated by the interplay of loss, sample placement, and optimization. 
A Gaussian will cause high image loss for two reasons: either its neighborhood is insufficiently sampled, or it has been erroneously placed.
When using loss for guidance, frequent densification can thus cause repeated duplication of misplaced Gaussians. 
However, when given sufficient time and observations, 3DGS will eliminate out-of-place Gaussians by lowering their opacity before densification occurs.

We invoke densification at a frequency of only one-fifth of 3DGS (i.e., every 500 iterations). Given a set of $N$ camera views, $V = \{ v_i \}_{i=1}^N$, the set of $M$ fitted Gaussians, $G = \{ g_j \}_{j=1}^M$, and the set of $N$ rendered views, $R = \{ r_i \}_{i=1}^N$, we evaluate a scoring function $F$ that is parameterized by per-Gaussian primitive attributes and projected per-pixel metrics. % values Gaussian contributes to, as well as its primitive attributes. 
This involves the following:
\begin{enumerate}
	\item \textbf{Determine per-view saliency matrix $\textbf{S}_{v}$}: For each view $v$, this matrix indicates pixels that may be undersampled (high loss) or contain high-frequency information. Additionally, this function enables prioritizing regions of interest:
	\begin{equation}
		\label{eq:pixel_importance}
		%i = \mathcal{P}(\mathit{l_1}, edges) \\
            \textbf{S}_{v} = \mathbbm{1}_{ROI}\odot(\lambda_1 \mathcal{L}_1(v, r_v) + \lambda_2 \mathit{E}(v)),\quad v \in V\\
	\end{equation}
	where $\mathcal{L}_1$ is the L1 loss, $E$ is a Laplacian filter, $\mathbbm{1}_{ROI}$ is a binary matrix indicating a masked region of interest, $\odot$ is the element-wise product, and $\lambda_1,\lambda_2$ are hyperparameters, set to 0.5 in our experiments.
	\item \textbf{Compute Gaussian scores $\textbf{S}_G$}:  We compute a global score vector $\textbf{S}_G$ that holds a score $S_g$ for each Gaussian $g$ in $G$. We do this by evaluating $F(\cdot)$ and summing over all $N$ views:

     \begin{align}
        \label{eq:gauss_score_1}
		%I = \mathcal{F}(\nabla, \mathcal{L}, \sigma, \mathcal{D}, O, r, d, \matcal{C})
  		S_{g} = \sum_{i}^{N}F(\nabla_{g}, 
    c^i_{g},
    \mathbbm{1}_g^i,
    \mathbf{D}^i_{g},
    \textbf{S}_{v}^i,  \mathbf{B}^i_{g},z^i_{g}, o_{g},  s_{g} )\\
         \textbf{S}_G = [S_{g_1}, ..., S_{g_M}]^T,\quad g_j \in G %\in  \mathbb{R}^M
    \end{align}
    Here, $\nabla_g$ is the Gaussian's positional gradient. $c^i_{g}$ denotes the number of pixels covered by $g$ in view $i$.  $\mathbbm{1}_g^i$ is a binary matrix that indicates these pixels. $\mathbf{D}_{g}^i$ is a matrix that holds the distance of each pixel to the center of $g$. 
    $\mathbf{B}_{g}^i$ contains each pixel's blending weight for $g$. Attributes $z^i_{g}$, $o_{g}$, and $s_{g}$ constitute the depth in $i$, opacity, and scale of $g$, respectively.
\end{enumerate}

$\textbf{S}_G$ is representative of the need to resample each Gaussian to converge to the final scene and serves as the foundation of our score-based densification. Alg.~\ref{algo:pseudocode} provides more details on this process. For the choice of $F$, we restrict each parameter's range using median scaling to remove outliers, followed by multiplication with $i$'s photometric loss. The so-rescaled parameters are then accumulated into a weighted sum, whose coefficients can be tuned for specific use cases. In the following, we explain the role of each parameter (and our proposed weighting) to achieve high quality with few Gaussians.

$\nabla_g$ (50): We adopt the magnitude of the positional gradient as a densification criterion from Kerbl et. al.~\cite{kerbl20233d}. According to the authors, high $\|\nabla_g\|_2$ can be interpreted as a 3D discontinuity detector. While provably effective, it alone usually leads to wasteful behavior and superfluous Gaussians. 

$c^i_g$ (0.1): The pixel count of $g$ acts as an indicator for primitives that tend to have large projections, which lead to a blurry appearance in rendered images.

$\textbf{D}^i_g$ (50): Splats that cover only a few pixels may still appear as thin elongated "slivers" on screen. We encourage their densification by scoring the accumulated distance of covered pixels to the center of $g$.

$\textbf{S}_{v}^i$ (10): We weight the accumulated per-pixel saliency scores of pixels covered by $g$ (i.e., the sum of element-wise products of $\textbf{S}_{v}^i$ and $\mathbbm{1}_g^i$). This enables the previously computed saliency to guide densification directly.

$\textbf{B}_{g}^i$ (50): The sum of per-pixel blending weights used in rendering indicates high-contributing Gaussians. Densifying them has the highest chance of causing visible changes in scene appearance and quality.

$z_{g}^i$ (5): The depth of each Gaussian allows us to distinguish between foreground and background. Note that this value is 0 for all Gaussians outside the view frustum: thus, it serves as a combined measurement of $g$'s visibility in the capture and its average distance to the camera. This prioritizes densifying commonly seen primitives without neglecting rarely seen background Gaussians.

$o_{g}$ (100): We use a high weight on opacity to steer densification away from low-opacity Gaussians. Low opacity is characteristic of floaters, or Gaussians that the optimization is currently phasing out.

$s_{g}$ (25): Overly large Gaussians---even if not observed up close during training---hurt generalizability to unseen views. Scoring the product of Gaussians' scales results in more uniformly sized primitives.

Given the final score vector $\textbf{S}_G$ and a budgeted target number $B$ of Gaussians to add, we perform densification by randomly resampling $B$ primitives from all Gaussians using $\textbf{S}_G$ as sampling weights.
In practice and for all experiments, we use $N=10$ uniformly sampled training views for computing per-pixel saliency scores. 
% \ssnote{10 views are empirically found to perform as good as including across all views, but with a significant reduction in time.}

\begin{algorithm}[t]
\caption{Proposed Steerable Densification Method}
\label{algo:pseudocode}
\setlength{\abovedisplayskip}{3pt}
\setlength{\belowdisplayskip}{3pt}
\begin{algorithmic}[1]
    \STATE T $\leftarrow$ Target Gaussian count at current iteration
    \STATE $\mathcal{G} \leftarrow$ All Gaussians $\{g_1, g_2, ..., g_{|G|}\}$
    \STATE $\mathcal{G}_t \leftarrow$ Gradient threshold
    \STATE $\mathcal{R}_t \leftarrow$ Radius threshold
    \FOR{image $i \in$ sampled views($N$)}
            \STATE $P_i \leftarrow$ Photometric loss
        %\STATE Pixel-wise saliency, $S_{v}$ from Eq. \ref{eq:pixel_importance}
        \STATE Initialise:  $c^i_g = 0; \mathbf{D}^i_g = 0; \textbf{s}^i_g = 0; \mathbf{B}^i_{g} = 0$
            \FOR{pixel $p$ $\in$ $i$} 
                \FOR{$g \in$ Gaussians contributing to $p$}
                    \STATE $c^i_g \mathrel{+}=1 $
                    \STATE 
              $\mathbf{D}^i_g \mathrel{+}=$ Distance from center of $g$ to $p$ 
    
                    \STATE $\textbf{s}^i_g \mathrel{+}=\textbf{S}_{v}^i(p)$
                    \STATE $\mathbf{B}^i_{g} \mathrel{+}=$ Blending weight of $g$ on $p$
                \ENDFOR
            \ENDFOR
            \STATE $S_{g} = S_{g} + P_i \cdot F(\nabla_{g},
                c^i_{g},
                \mathbf{D}^i_{g},
                \textbf{s}_{g}^i,  \mathbf{B}^i_{g},z^i_{g}, o_{g}, s_{g})$
        \ENDFOR
        \STATE $\textbf{S}_G = [S_{g_1}, ..., S_{g_M}]^T,\quad g_i \in \mathcal{G}$
        \STATE $B = T-|\mathcal{G}|$ \hfill $\rhd$ \#Gaussians to add
       
    \STATE Top gaussian indices: $G' \sim (\mathcal{G}, \textbf{S}_G, B)$
    %\FOR{$g$ in $\mathcal{G}$}
        \FOR{$g \in G'$}
            \IF{($\nabla$\textsubscript{g} $>$ $\mathcal{G}_t$) $\And$ (radius\textsubscript{g} $>$ $\mathcal{R}_t$)}
            \STATE SPLIT
            \ELSIF{($\nabla$\textsubscript{g} $>$ $\mathcal{G}_t$) $\And$ (radius\textsubscript{g} $\leq$ $\mathcal{R}_t$)}
            \STATE CLONE
        \ENDIF
        %\ENDIF
    \ENDFOR
\end{algorithmic}
\end{algorithm}

\subsection{High-Opacity Gaussians}
While the basic Gaussian primitives of 3DGS can yield high quality, their expressiveness is limited by their rigid Gaussian falloff~\cite{hamdi_2024_CVPR}. To remedy this, Kerbl et al.~\cite{hierarchicalgaussians24} used simple clamped Gaussians with opacities ${>}1$ to approximate the appearance of Gaussian clusters in a hierarchical level-of-detail structure. We find that these high-opacity Gaussians can also boost the ability for modeling opaque surfaces with a low number of primitives. Therefore, we convert the regular, capped Gaussian primitives to high-opacity Gaussians after reaching the midpoint of our training (15K iterations). This involves replacing the opacity activation with \texttt{abs} and clamping blending weights to $1$ from above during rendering. As shown by our ablation, this change positively impacts quality metrics, particularly PNSR.

\section{3DGS Runtime Analysis and Optimization}

\begin{figure}
\begin{subfigure}{0.59\textwidth}
\includegraphics[width=\columnwidth]{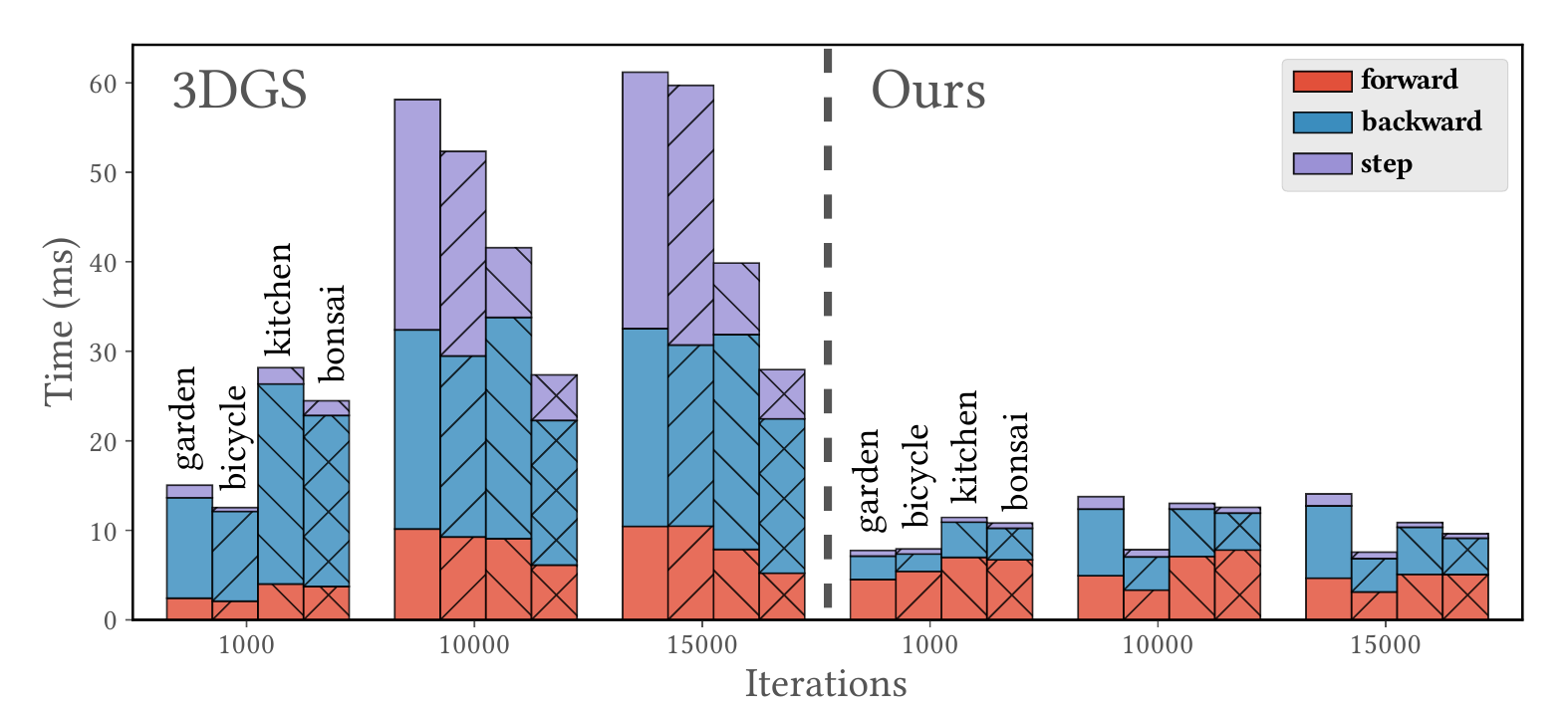}
\caption{Iteration Time Breakdown}
% First four clusters are 3DGS. Next 4 clusters is us. In each cluster there are 4 scenes: garden, bicycle, kitchen, bonsai. RTX 4090. "Ours" is with Saswat's optimizer.\bknote{check caption}} \ssnote{Would it be neater if the first and second 4 clusters were separated by a different colored background?}
\label{fig:runtime_analysis}
\end{subfigure}
\hfill
\begin{subfigure}{0.39\columnwidth}
    \includegraphics[width=\columnwidth]{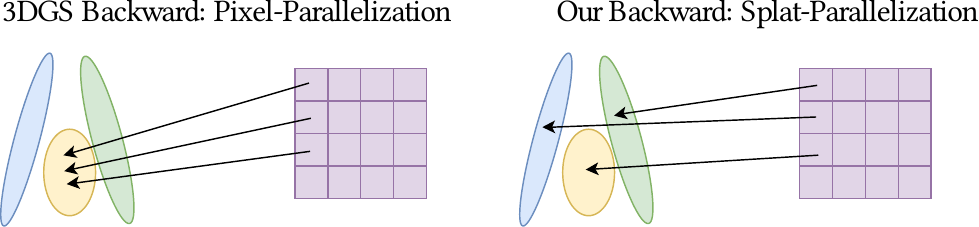}
    \\
    \vspace{0.4cm}
    \includegraphics[width=\columnwidth]{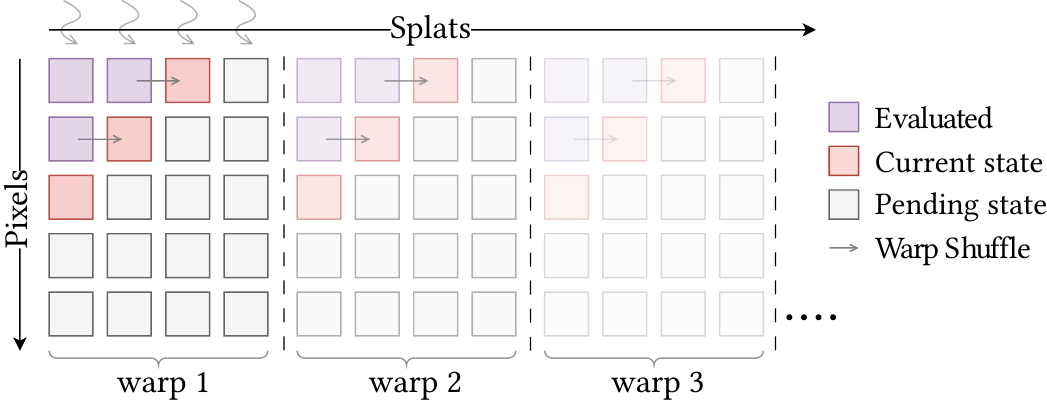}
    \caption{Per-Gaussian Backward %This procedure enables a very high degree of splat-parallelism and avoids traversal over a sorted list. \bknote{check caption}
    }
    \label{fig:backward}
\end{subfigure}
\caption{(a) Time spent in different parts (forward pass, backward pass, optimizer step) of one 3DGS iteration in four scenes (\textsc{garden}, \textsc{bicycle}, \textsc{kitchen}, \textsc{bonsai}). Left: analysis of original 3DGS at different stages of training. Right: using our budgeted densification and performance optimizations. (b) Gradient backpropagation. (Top) 3DGS utilizes per-pixel parallelization for backpropagation. Atomic gradient additions create frequent collisions, slowing down the backward. Instead, we parallelize on the projected 2D splats, such that each thread (and pixel) contributes to one Gaussian at a time. (Bottom)~The gradient calculation requires processing a set of per-pixel, per-splat values resulting in an implicit traversal of a splat~$\iff$~pixel state table. During the forward, we store the pixel states for every $32^{\text{nd}}$ splat in the sorted list. For the backward, we divide the splats into buckets of size $32$, each of which gets scheduled to a CUDA warp. Warps use intra-warp shuffling to produce their share of the state table cheaply.}
\end{figure}

To better understand the performance challenges of 3DGS, we benchmark the original training pipeline, written in PyTorch, with explicit CUDA extensions for differentiable rasterization. We provide a breakdown of the time taken by the high-level steps in each iteration for multiple scenes, at different stages of training, in Fig.~\ref{fig:runtime_analysis}. We note that, throughout the training routine, backpropagation of gradients is the dominating bottleneck, closely followed by \textsc{Adam} optimizer updates as the number of Gaussians increases. With these insights, we propose targeted solutions for accelerating 3DGS training.

\subsection{Backpropagation with Per-Splat Parallelization}
\begin{figure}
\floatbox[{\capbeside\thisfloatsetup{capbesideposition={right,center},capbesidewidth=5cm}}]{figure}[\FBwidth]
{\caption{Backward pass duration in training of \textsc{Bicycle} using 3DGS, DISTWAR~\cite{distwar} and our variants. For our approach, we plot the times when used with original 3DGS densification and with our more tightly budgeted schedule.}\label{fig:backward_comp}}
{\includegraphics[width=10cm]{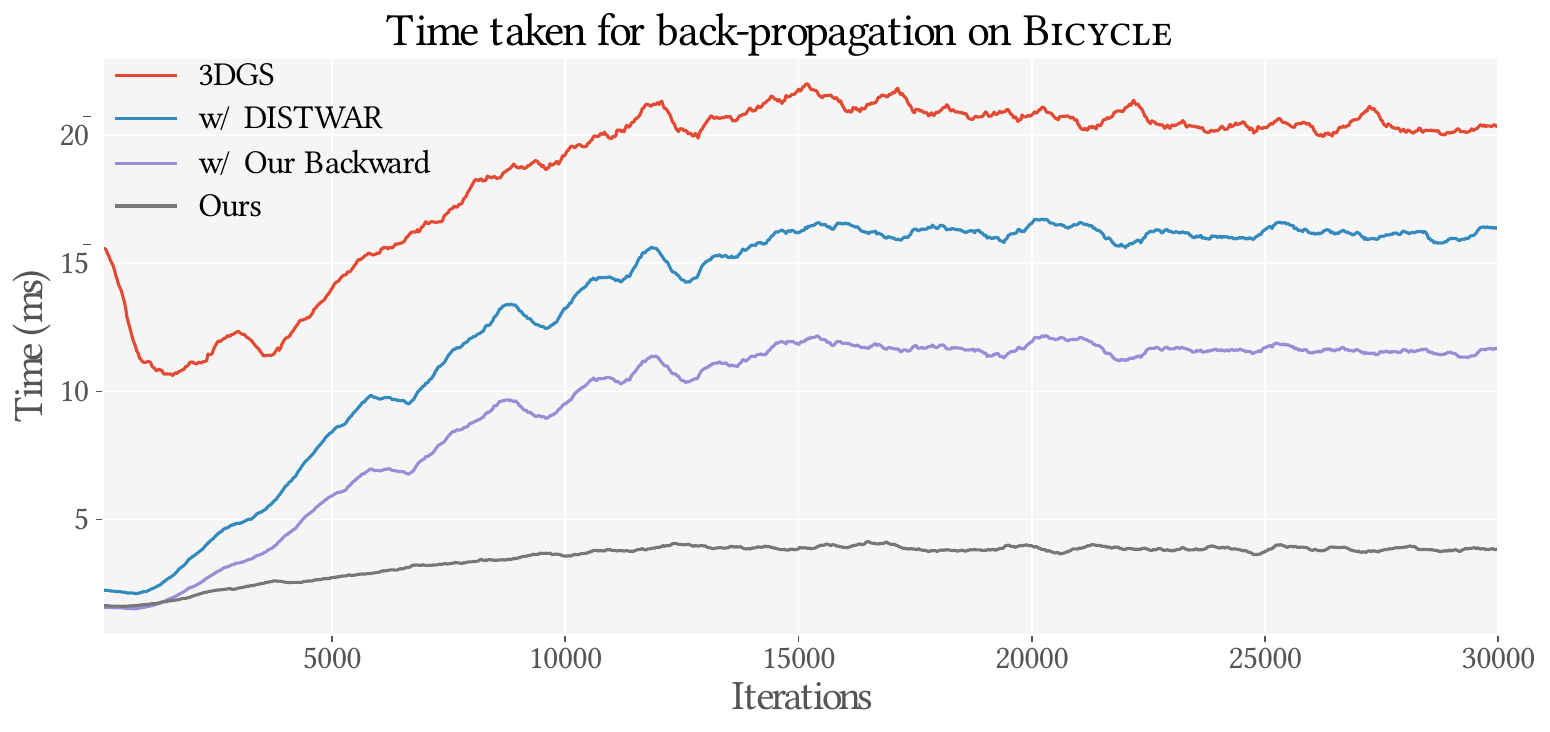}}
\end{figure}

{
In the original 3DGS backward pass, gradients are propagated from the pixels onto the Gaussians. The total gradient calculation involves computing many per-pixel, per-splat values, which are then accumulated globally via reduction. Kerbl et al.\ \cite{kerbl20233d} take the natural approach of mapping threads to pixels and iterating over the depth-sorted splats back-to-front. Within a tile, each thread considers splats in reverse blending order, evaluates a per-pixel gradient portion, and atomically adds it to the corresponding splat's accumulated gradient. While correct, this leads to multiple threads contending for access to the same locations and thus serialized atomic operations, as shown in Fig.~\ref{fig:backward}.
The fact that each Gaussian splat maintains \emph{a multitude} of gradients for its attributes further aggravates the overhead of this reduction~\cite{distwar}.

We propose a solution where each tile uses a parallelization scheme over the 2D \emph{splats} instead of pixels. This new approach lets threads maintain a per-splat state and continually exchange per-pixel states consisting of transmittance $T$ and accumulated color $RGB$ (as opposed to storing per-pixel information and exchanging the larger per-splat data). %every splat iterating over the pixels in the tile to accumulate their gradient contributions, eliminating the repeated usage of atomics completely.
Ignoring corner cases, let us assume a simplified setting where \mbox{\#threads $=$ \#pixels $=$ \#splats $=$ N}.
At each point in time, thread $i$ computes a gradient portion for splat $i$; to do this, it requires the state of each pixel $j$ after blending the frontmost $i$ primitives.
During the forward pass, each thread stores one per-pixel state every N splats in the autodiff context for backward, resulting in available starting states $(0, j), (N, j), ... \forall j$. From these, each thread in a tile generates pixel state $(i, j)$ at the beginning of the backward pass. Threads then exchange pixel states via fast collaborative sharing. In each step, thread $i+1$ applies the default alpha blending logic to go from its received $(i,j)$ to $(i+1,j)$ and incorporates this information into the gradient.
For more details please refer to Fig. \ref{fig:backward}.
%and accompanying video. 

We also observe that iterating the tail of each tile's depth-sorted list of splats often becomes redundant due to occlusion. This is avoided in the forward pass, which terminates upon saturation. To exploit this in backpropagation as well, we keep track of the last contributor across the tile and use it to skip entire groups of splat$\iff$tile pairings.
Finally, we reduce the overall rasterization workload via tighter culling as proposed by Radl et al.~\cite{radl2024stopthepop}, minimizing redundant splats in the forward and backward pass.

Fig.~\ref{fig:backward_comp} compares the time taken for the backward methods of 3DGS, concurrent work DISTWAR~\cite{distwar}, and Ours both with the original 3DGS and our budgeted optimization schedule.

\subsection{Accelerated SH and Differentiable Loss Computation}
Fig.~\ref{fig:runtime_analysis} reveals the significant time spent on \textsc{Adam} updates as the number of Gaussians increases. Of these updates, SHs---48 out of 59 optimized per-Gaussian attributes---are responsible for the vast majority. To amend this, we switch all bands beyond the first to a batched update schedule, performing only one step of \textsc{Adam} optimization every 16 iterations. 
The original 3DGS implementation combines the 0th SH band (i.e., base color) and higher bands into a single tensor before rasterization. This consumes a surprising portion of the forward pass. We avoid this by extending the differential rasterizer to load Gaussian SH coefficients from separate tensors.

3DGS loss computation involves evaluating the SSIM metric. It is configured to use 11$\times$11 Gaussian kernel convolution: we propose using optimized CUDA kernels to perform differentiable 2D convolution via two consecutive 1D convolutions since Gaussian kernels are separable in nature. In addition, we use a fused kernel for the evaluation of the SSIM metric from the convolved results. This speeds up the loss calculation and is particularly impactful when the number of optimized Gaussians is low compared to image resolution, which is the case when training on a budget.

These measures---along with our per-splat backward---serve as drop-in replacements to the original 3DGS. The mentioned changes, except for the modified SH update schedule, yield equivalent results to 3DGS. Our experiments show that these considerably reduce 3DGS training times and can outperform the recently released version 1.0 of gsplat~\cite{ye2023mathematical} by $1.5\times$--$2\times$. We provide the source code for our optimizations at \url{https://github.com/nullptr81/3dgs-accel}.

%Finally, we modify the differentiable rasterizer to take the direct color (degree 0 SH) and remaining SH coefficients separately since they are have different learning rates. This bypasses a heavy tensor concatenation step which also holds back the training speeds.
}

% \begin{figure}
% \centering
% \includegraphics[width=0.69\columnwidth]{assets/backward_comp.pdf}
% \caption{Backward pass duration in training of \textsc{Bicycle} using 3DGS, DISTWAR~\cite{distwar} and our variants. For our approach, we plot the times when used with original 3DGS densification and with our own.
% \label{fig:backward_comp}
% }
% \end{figure}

\section{Evaluation and Discussion}
\label{sec:eval}
%\subsection{Implementation details}
This section evaluates our proposed approach both quantitatively and qualitatively. Our implementation is based on top of the original 3DGS codebase~\cite{kerbl20233d}. 
Most original hyperparameters are retained; however, we added a separate $\textsc{Adam}$ optimizer for batched SH updates, increasing the SH learning rate four times ($0.001$) and reducing the opacity learning rate by half ($0.025$).  The evaluation was conducted using an NVIDIA RTX A4500 GPU. Results for other techniques, including training times, were obtained on the same hardware or adjusted to ensure comparability.

\subsection{Datasets and Metrics}
We run benchmarks on three established datasets---Tanks\&Temples \cite{Knapitsch2017}, Deep Blending~\cite{DeepBlending2018}, and MipNeRF360~\cite{barron2022mipnerf360}, which contain 2, 2, and 9 scenes, respectively. These datasets cover bounded indoor and unbounded outdoor scenarios with detailed backgrounds. We use the same train/test split as the original 3DGS publication and follow-up work. 

In addition to common quality metrics (peak signal-to-noise ratio (PSNR), structural similarity (SSIM), and perceptual similarity (LPIPS)~\cite{zhang2018perceptual}, an important focus of our work is resource efficiency: Our method aims to achieve high quality with low resource usage. We assess these qualities by timing the optimization (Train time), counting the final number of Gaussians (\#G), as well as recording the \emph{peak} number (Peak \#G) during training. 

\begin{table}
\caption{Quantitative comparison with other methods in two scenarios (top half \& bottom half). For quality, we compare PSNR, SSIM, and LPIPS for quality. For resource efficiency, we report training time, and, where applicable, the final number (\#G), and peak number (Peak \#G) of Gaussians used. \textbf{Best} and \underline{second-best} results are highlighted for each.}
\label{tab:results}

\scalebox{0.67}{

\begin{tabular}{l|c|c|c|c|c|c|c|c|c|c|c|c|c|c|c|c|c|c|}
\multicolumn{1}{c}{} & \multicolumn{6}{c|}{Tanks\&Temples}                                                                                                                                                                                                                                                                                       & \multicolumn{6}{c|}{MipNeRF-360}                                                                                                                                                                                                                                                                                                               & \multicolumn{6}{c}{Deep Blending}                                                                                                                                                                                                                                                                                                            \\ \cline{2-19} 
                  & \multicolumn{1}{c|}{SSIM}           & \multicolumn{1}{c|}{PSNR}           & \multicolumn{1}{c|}{LPIPS}          & \multicolumn{1}{c|}{\begin{tabular}[c]{@{}c@{}}Train\\ time\end{tabular}} & \multicolumn{1}{c|}{\begin{tabular}[c]{@{}c@{}}\#G \\ ($10^6$)\end{tabular}} & \multicolumn{1}{c|}{\begin{tabular}[c]{@{}c@{}}Peak \\ \#G\end{tabular}}           & \multicolumn{1}{c|}{SSIM}           & \multicolumn{1}{c|}{PSNR}           & \multicolumn{1}{c|}{LPIPS}          & \multicolumn{1}{c|}{\begin{tabular}[c]{@{}c@{}}Train\\ time\end{tabular}} & \multicolumn{1}{c|}{\begin{tabular}[c]{@{}c@{}}\#G \\ ($10^6$)\end{tabular}} & \multicolumn{1}{c|}{\begin{tabular}[c]{@{}c@{}}Peak \\ \#G\end{tabular}}             & \multicolumn{1}{c|}{SSIM}          & \multicolumn{1}{c|}{PSNR}           & \multicolumn{1}{c|}{LPIPS}          & \multicolumn{1}{c|}{\begin{tabular}[c]{@{}c@{}}Train\\ time\end{tabular}} & \multicolumn{1}{c|}{\begin{tabular}[c]{@{}c@{}}\#G \\ ($10^6$)\end{tabular}} & \multicolumn{1}{c|}{\begin{tabular}[c]{@{}c@{}}Peak \\ \#G\end{tabular}}

 \\
 \hline  

C3DGS             & \multicolumn{1}{c|}{0.843}          & \multicolumn{1}{c|}{23.57}          & \multicolumn{1}{c|}{0.182}          & \multicolumn{1}{c|}{28\,m}                                                   & \multicolumn{1}{c|}{1.53}                                                 & \multicolumn{1}{c|}{1.84}          &\multicolumn{1}{c|}{0.811}          & \multicolumn{1}{c|}{\underline{27.34}}    & \multicolumn{1}{c|}{0.221}          & \multicolumn{1}{c|}{43\,m}                                                   & \multicolumn{1}{c|}{2.44}                                                 & \multicolumn{1}{c|}{2.94}          &\multicolumn{1}{c|}{0.900}           & \multicolumn{1}{c|}{29.54} & 0.252          & \multicolumn{1}{c|}{39\,m}                                                   & \multicolumn{1}{c|}{2.43}                                                 & \multicolumn{1}{c|}{2.81}                        \\
RVQ               & \multicolumn{1}{c|}{0.831}          & \multicolumn{1}{c|}{23.30}           & \multicolumn{1}{c|}{0.202}          & \multicolumn{1}{c|}{27\,m}                                                   & \multicolumn{1}{c|}{0.83}                                                  & \multicolumn{1}{c|}{1.46}          & \multicolumn{1}{c|}{0.797}          & \multicolumn{1}{c|}{26.99}          & \multicolumn{1}{c|}{0.245}          & \multicolumn{1}{c|}{48\,m}                                                   & \multicolumn{1}{c|}{1.41}                                                 & \multicolumn{1}{c|}{2.57}                              & \multicolumn{1}{c|}{0.901}         & \multicolumn{1}{c|}{29.75}          & \multicolumn{1}{c|}{0.260}           & \multicolumn{1}{c|}{38\,m}                                                   & \multicolumn{1}{c|}{1.04}                                                 & \multicolumn{1}{c|}{2.25}   \\
\cite{papanto2024}  & \multicolumn{1}{c|}{\underline{0.844}}    & \multicolumn{1}{c|}{\underline{23.66}}    & \multicolumn{1}{c|}{\textbf{0.178}} & \multicolumn{1}{c|}{\underline{18\,m}}                                                     & \multicolumn{1}{c|}{0.71}                                                            & \multicolumn{1}{c|}{\underline{0.71}}              &  \multicolumn{1}{c|}{\underline{0.814}}    & \multicolumn{1}{c|}{\textbf{27.43}} & \multicolumn{1}{c|}{\underline{0.220}}    & \multicolumn{1}{c|}{25\,m}                                                     & \multicolumn{1}{c|}{0.83}                                                            & \multicolumn{1}{c|}{\underline{0.83}}              & \multicolumn{1}{c|}{0.902}         & \multicolumn{1}{c|}{29.57}          & \multicolumn{1}{c|}{\underline{0.247}}    & \multicolumn{1}{c|}{22\,m}                                                     & \multicolumn{1}{c|}{0.97}                                                            & \multicolumn{1}{c|}{\underline{0.97}}                                 \\
Mini-Splatting    & \multicolumn{1}{c|}{\textbf{0.847}} & \multicolumn{1}{c|}{23.42}          & \multicolumn{1}{c|}{\underline{0.181}}    & \multicolumn{1}{c|}{20\,m}                                             & \multicolumn{1}{c|}{\underline{0.31}}                                                  & \multicolumn{1}{c|}{4.32}          & \multicolumn{1}{c|}{\textbf{0.822}} & \multicolumn{1}{c|}{27.26}          & \multicolumn{1}{c|}{\textbf{0.217}} & \multicolumn{1}{c|}{30\,m}                                             & \multicolumn{1}{c|}{\textbf{0.50}}                                                  & \multicolumn{1}{c|}{4.32} & \multicolumn{1}{c|}{\textbf{0.909}} & \multicolumn{1}{c|}{\textbf{30.04}} & \multicolumn{1}{c|}{\textbf{0.244}} & \multicolumn{1}{c|}{24\,m}                                             & \multicolumn{1}{c|}{\underline{0.56}}                                                       & \multicolumn{1}{c|}{4.51}                         \\
INGP-Big          & \multicolumn{1}{c|}{0.745}          & \multicolumn{1}{c|}{21.92}          & \multicolumn{1}{c|}{0.305}          & \multicolumn{1}{c|}{\textbf{7\,m}}                                                 & \multicolumn{1}{c|}{-}                                                            & \multicolumn{1}{c|}{-}              & \multicolumn{1}{c|}{0.699}          & \multicolumn{1}{c|}{25.59}          & \multicolumn{1}{c|}{0.331}          & \multicolumn{1}{c|}{\textbf{8\,m}}                                                 & \multicolumn{1}{c|}{-}                                                            & \multicolumn{1}{c|}{-}                 & \multicolumn{1}{c|}{0.817}         & \multicolumn{1}{c|}{24.96}           & \multicolumn{1}{c|}{0.390}          & \multicolumn{1}{c|}{\underline{8\,m}}                                                 & \multicolumn{1}{c|}{-}                                                            & \multicolumn{1}{c|}{-}                         \\
Ours             & \multicolumn{1}{c|}{0.837}          & \multicolumn{1}{c|}{\textbf{23.95}} & \multicolumn{1}{c|}{0.201}          & \multicolumn{1}{c|}{\textbf{7\,m}}                                           & \multicolumn{1}{c|}{\textbf{0.29}}                                         & \multicolumn{1}{c|}{\textbf{0.29}} & \multicolumn{1}{c|}{0.801}          & \multicolumn{1}{c|}{27.31}          & \multicolumn{1}{c|}{0.252}          & \multicolumn{1}{c|}{\underline{11\,m}}                                          & \multicolumn{1}{c|}{\underline{0.63}}                                         & \multicolumn{1}{c|}{\textbf{0.63}}    & \multicolumn{1}{c|}{\underline{0.904}}   & \multicolumn{1}{c|}{\underline{29.82}}    & \multicolumn{1}{c|}{0.260}           & \multicolumn{1}{c|}{\textbf{7\,m}}                                           & \multicolumn{1}{c|}{\textbf{0.27}}                                         & \multicolumn{1}{c|}{\textbf{0.27}}
                  
                  \\
                  \hline
Plenoxels & 0.719 & 21.08 & 0.379 & \underline{25\,m} & -  & -  & 0.626 & 23.08 & 0.463 & \textbf{26\,m} & - & -  & 0.795 & 23.06 & 0.51 & \underline{28\,m} & -  & - \\
MipNeRF360          & \multicolumn{1}{c|}{0.759}          & \multicolumn{1}{c|}{22.22}          & \multicolumn{1}{c|}{0.257}          & \multicolumn{1}{c|}{48\,h}                                                 & \multicolumn{1}{c|}{-}                                                            & \multicolumn{1}{c|}{-}              & \multicolumn{1}{c|}{0.792}          & \multicolumn{1}{c|}{27.69}          & \multicolumn{1}{c|}{0.237}          & \multicolumn{1}{c|}{48\,h}                                                 & \multicolumn{1}{c|}{-}                                                            & \multicolumn{1}{c|}{-}                 & \multicolumn{1}{c|}{0.901}         & \multicolumn{1}{c|}{29.4}           & \multicolumn{1}{c|}{0.245}          & \multicolumn{1}{c|}{48\,h}                                                 & \multicolumn{1}{c|}{-}                                                            & \multicolumn{1}{c|}{-}                         \\
Zip-NeRF          & \multicolumn{1}{c|}{-}               & \multicolumn{1}{c|}{-}               & \multicolumn{1}{c|}{-}               & \multicolumn{1}{c|}{-}                                                     & \multicolumn{1}{c|}{-}                                                            & \multicolumn{1}{c|}{-}              & \multicolumn{1}{c|}{\textbf{0.828}}          & \multicolumn{1}{c|}{\textbf{28.54}}          & \multicolumn{1}{c|}{\textbf{0.189}}          & \multicolumn{1}{c|}{1.5\,h}                                                     & \multicolumn{1}{c|}{-}                                                            & \multicolumn{1}{c|}{-}              & \multicolumn{1}{c|}{-}    & \multicolumn{1}{c|}{-}              & \multicolumn{1}{c|}{-}              &                                                    - & \multicolumn{1}{c|}{-}                                                            & \multicolumn{1}{c|}{-}                        \\
3DGS              & \multicolumn{1}{c|}{\underline{0.847}}          & \multicolumn{1}{c|}{\underline{23.65}}          & \multicolumn{1}{c|}{\underline{0.176}}          & \multicolumn{1}{c|}{28\,m}                                                   & \multicolumn{1}{c|}{1.84}                                                 & \multicolumn{1}{c|}{1.84}          & \multicolumn{1}{c|}{0.815}          & \multicolumn{1}{c|}{27.46}          & \multicolumn{1}{c|}{0.215}          & \multicolumn{1}{c|}{43\,m}                                                   & \multicolumn{1}{c|}{3.31}                                                 & \multicolumn{1}{c|}{3.31}          &                          \multicolumn{1}{c|}{\underline{0.904}}         & \multicolumn{1}{c|}{\underline{29.64}}          & \multicolumn{1}{c|}{\underline{0.243}}          & \multicolumn{1}{c|}{39\,m}                                                   & \multicolumn{1}{c|}{2.81}                                                 & \multicolumn{1}{c|}{2.81}           \\
Ours (\#3DGS) & \multicolumn{1}{c|}{\textbf{0.851}}          & \multicolumn{1}{c|}{\textbf{24.04}} & \multicolumn{1}{c|}{\textbf{0.170}}          & \multicolumn{1}{c|}{\textbf{20\,m}}                                           & \multicolumn{1}{c|}{1.84}                                         & \multicolumn{1}{c|}{1.84} & \multicolumn{1}{c|}{\underline{0.822}}          & \multicolumn{1}{c|}{\underline{27.79}}          & \multicolumn{1}{c|}{\underline{0.205}}          & \multicolumn{1}{c|}{\underline{32\,m}}                                          & \multicolumn{1}{c|}{3.31}                                         & \multicolumn{1}{c|}{3.31}    & \multicolumn{1}{c|}{\textbf{0.907}}   & \multicolumn{1}{c|}{\textbf{30.14}}    & \multicolumn{1}{c|}{\textbf{0.235}}           & \multicolumn{1}{c|}{\textbf{22\,m}}                                           & \multicolumn{1}{c|}{2.81}                                         & \multicolumn{1}{c|}{2.81} \\ 

\hline
                     
\end{tabular}
}
\end{table}

\pagebreak
	
\subsection{Results}
We evaluate our method in two separate, budgeted scenarios (top/bottom of Table \ref{tab:results}). %Results for the first scenario are shown at the top of Table \ref{tab:results}, and those for the second scenario at the bottom. 
For qualitative results, see Fig.~\ref{fig:fig1}.

In the first, we select a reasonable budget for individual scenes, based on their spatial extent and SfM point count. For the small-scale indoor scenes in MipNeRF-360, we set the budget to $2\times$ the SfM points. For the larger, full-room indoor captures of Deep Blending, we use $5\times$, and for unbounded outdoor scenes, we use $15\times$. For the outdoor Tanks\&Temples, the initial SfM point count is significantly higher, thus we set the budget to $2\times$ here as well. Note that this parameterization could be automatized by providing scenes in real-world coordinates or a corresponding multiplier. To evaluate the resources/quality tradeoff, we compare with recent works that aim at reducing the memory footprint of 3DGS: (Compressed 3DGS)~\cite{niedermayr2023compressed}, Compact-3DGS (R-VQ)~\cite{lee2023compact}, and \cite{papanto2024}. 
Due to its exceptionally fast training, we also compare with the high-quality version of Instant-NGP (INGP-Big)~\cite{muller2022instant}. To perform a thorough evaluation and provide comprehensive context, we also evaluate the concurrent pre-print for Mini-Splatting~\cite{fang2024minisplatting}. 
Assessing the results in the top half of Table \ref{tab:results}, we find that among splatting-based methods, Ours achieves outstanding reduction (slightly outperformed only by Mini-Splatting in one dataset). Notably, our budgeted method is competitive with (and sometimes surpasses) 3DGS in terms of quality, especially PSNR. However, the most striking benefit of our approach is efficiency: Mini-Splatting---similar to 3DGS---relies on heavily oversampling the scene before pruning, creating a vast gap of up to $10\times$ between their peak and final model size. In contrast, our method uses a purely constructive optimization that only adds Gaussians towards an exact target budget. In addition, we achieve this using between half and one-third of the time of the next-fastest 3DGS-based methods and occasionally outperform even Instant-NGP in terms of speed. 

In the second budgeted scenario, we configure our optimization to reach the exact same model size as the original 3DGS. Since the expressiveness of our method rises with the available budget, in this scenario, we compare our results with representative, high-quality approaches from different domains: Plenoxels~\cite{Fridovich-Keil_2022_CVPR}, and two sophisticated NeRF methods, MipNeRF360~\cite{barron2022mipnerf360} and ZipNeRF~\cite{barron2023zipnerf}. Finally, we consider the original 3DGS technique \cite{kerbl20233d}. We provide the corresponding results in the bottom half of Table \ref{tab:results}.
Although our optimization differs significantly from 3DGS, we demonstrate that our budgeting mechanism allows to match their model size exactly. The achieved quality easily surpasses 3DGS and MipNeRF360, second only to the recent, much slower Zip-NeRF approach.

\newcommand{\spyimg}[4]{%
	\begin{tikzpicture}[spy using outlines={yellow,magnification=3,size=0.95cm, connect spies}]
		\node[anchor=south west,inner sep=0] at (0,0) {\includegraphics[width=#1]{#2}};
		\spy on (#3) in node [left] at (#4);
	\end{tikzpicture}%
}

\newcommand{\spyimgred}[4]{%
\begin{tikzpicture}[spy using outlines={red,magnification=3,size=0.95cm, connect spies}]
\node[anchor=south west,inner sep=0] at (0,0) {\includegraphics[width=#1]{#2}};
\spy on (#3) in node [left] at (#4);
\end{tikzpicture}%
}

\begin{figure*}[!h]
   \centering
   \setlength{\tabcolsep}{1pt}
\renewcommand{\arraystretch}{0.5}
\begin{tabular}{ccccc}
    {\small Ground Truth} & {\small 3DGS} & {\small INGP-Big} & {\small Ours} & {\small Ours (\#3DGS)}\\

    \spyimg{0.19\textwidth}{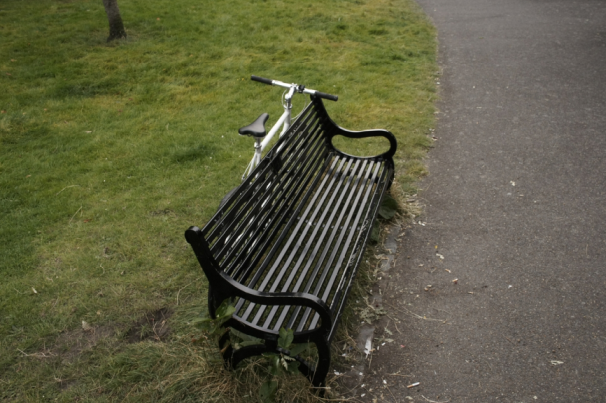}{2.8,1.8}{3,0.7}&
    \spyimg{0.19\textwidth}{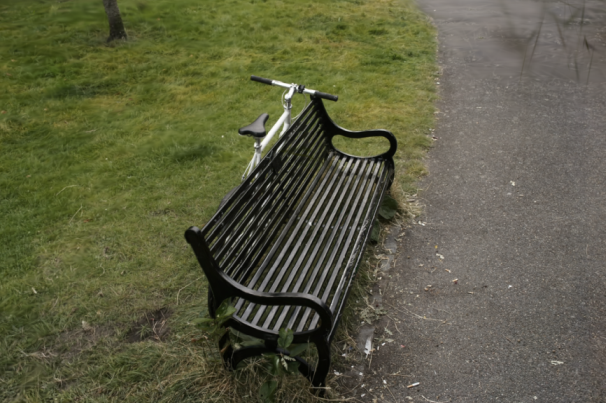}{2.8,1.8}{3,0.7}&
    \spyimg{0.19\textwidth}{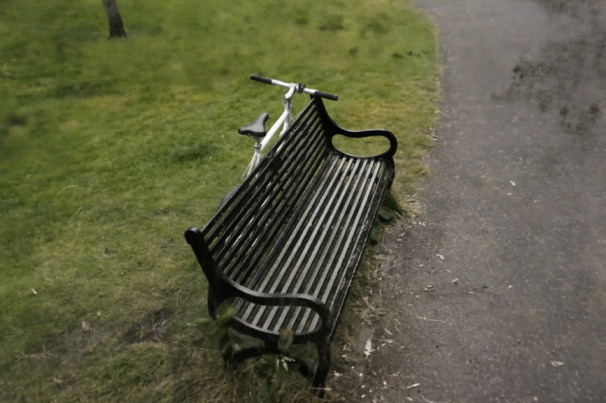}{2.8,1.8}{3,0.7}&
    \spyimg{0.19\textwidth}{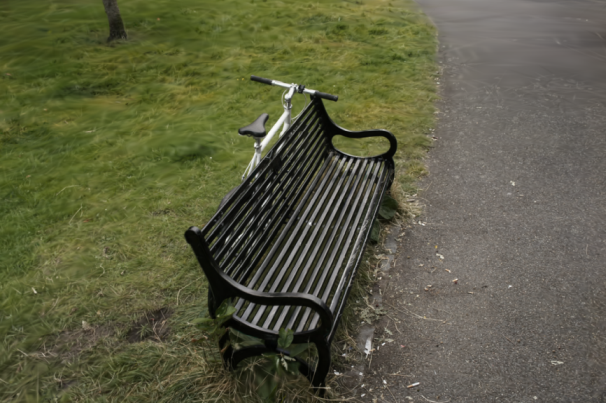}{2.8,1.8}{3,0.7}&
    \spyimg{0.19\textwidth}{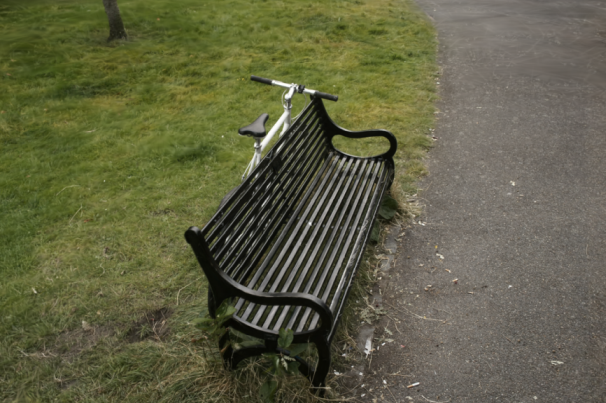}{2.8,1.8}{3,0.7}\\

    \spyimgred{0.19\textwidth}{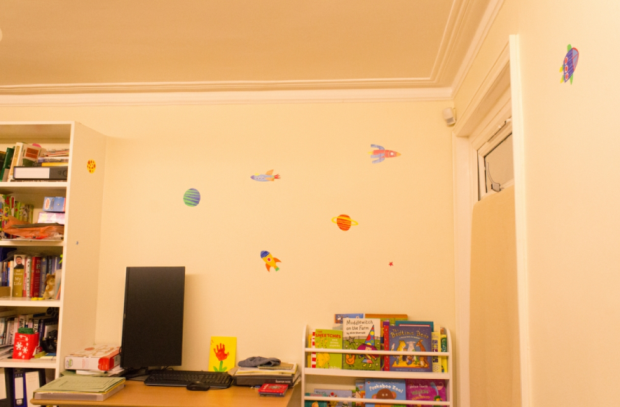}{2.9,1.75}{3,0.7}&
    \spyimgred{0.19\textwidth}{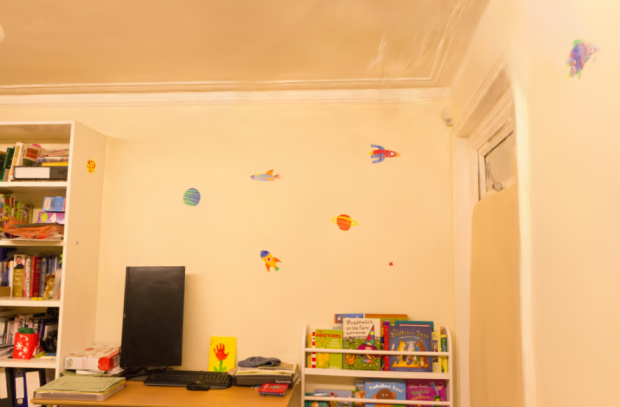}{2.9,1.75}{3,0.7}&
    \spyimgred{0.19\textwidth}{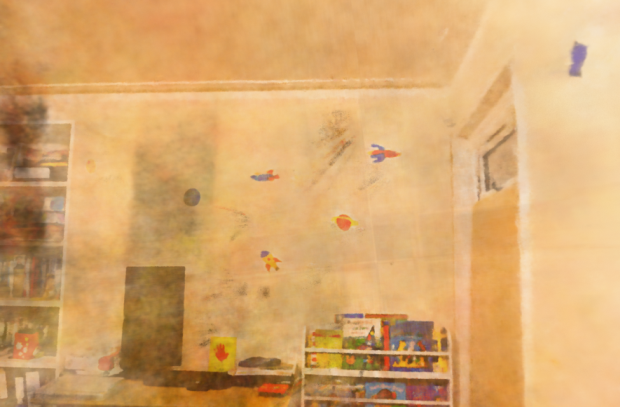}{2.9,1.75}{3,0.7}&
    \spyimgred{0.19\textwidth}{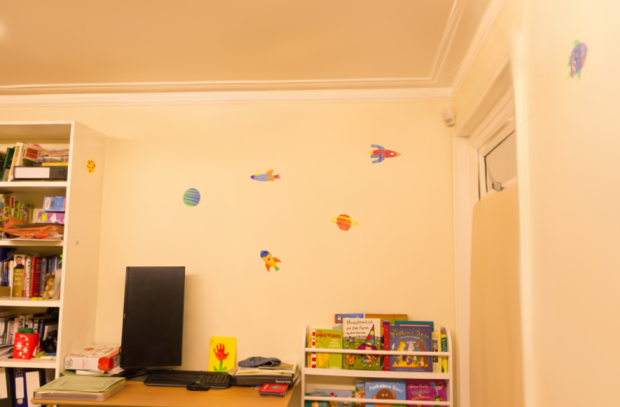}{2.9,1.75}{3,0.7}&
    \spyimgred{0.19\textwidth}{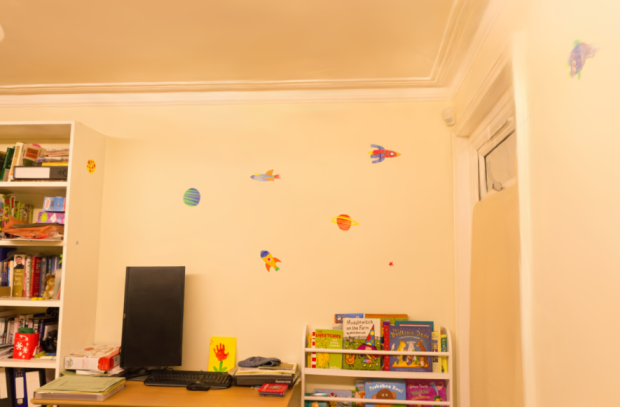}{2.9,1.75}{3,0.7}\\

    \spyimg{0.19\textwidth}{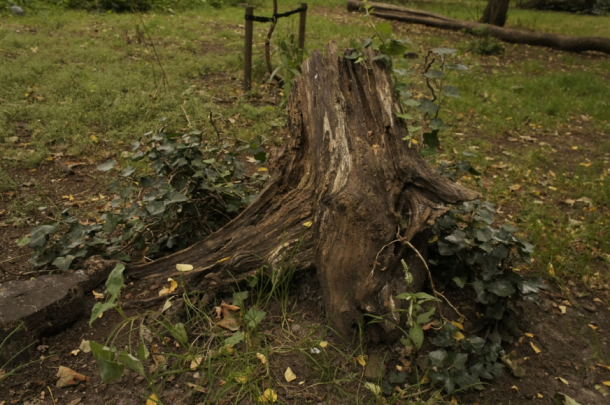}{2.9,1.75}{3,0.7}&
    \spyimg{0.19\textwidth}{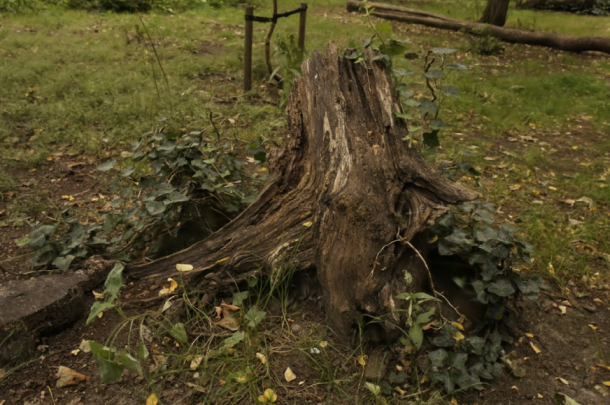}{2.9,1.75}{3,0.7}&
    \spyimg{0.19\textwidth}{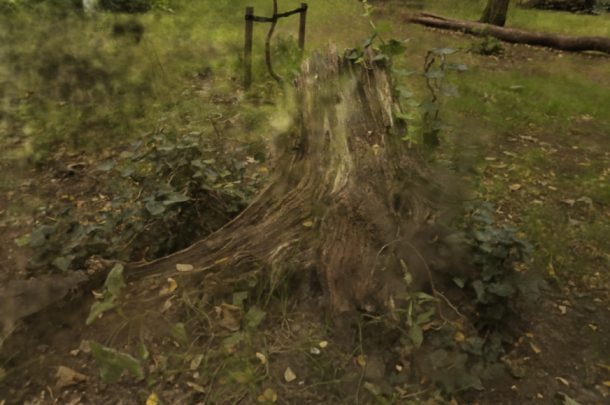}{2.9,1.75}{3,0.7}&
    \spyimg{0.19\textwidth}{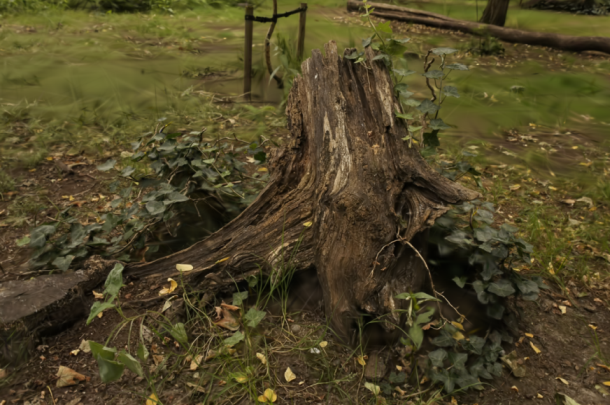}{2.9,1.75}{3,0.7}&
    \spyimg{0.19\textwidth}{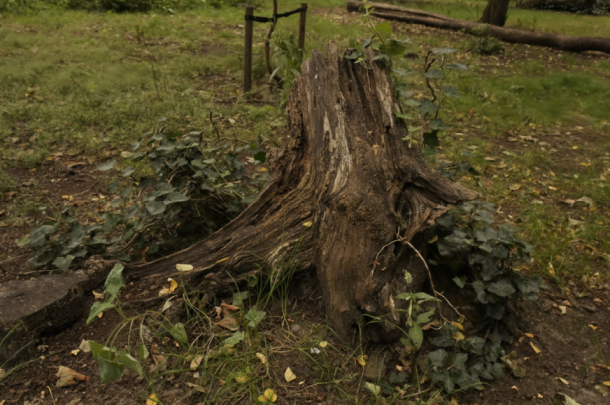}{2.9,1.75}{3,0.7} \\
    
    \spyimg{0.19\textwidth}{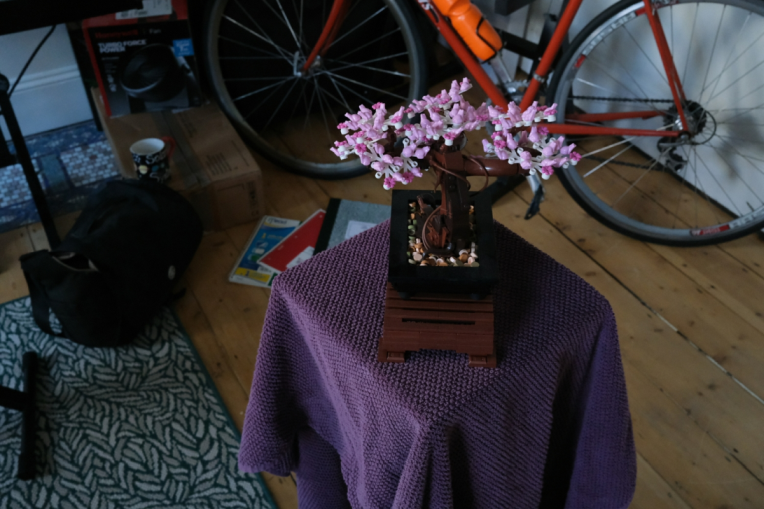}{2.9,1.75}{3,0.7}&
    \spyimg{0.19\textwidth}{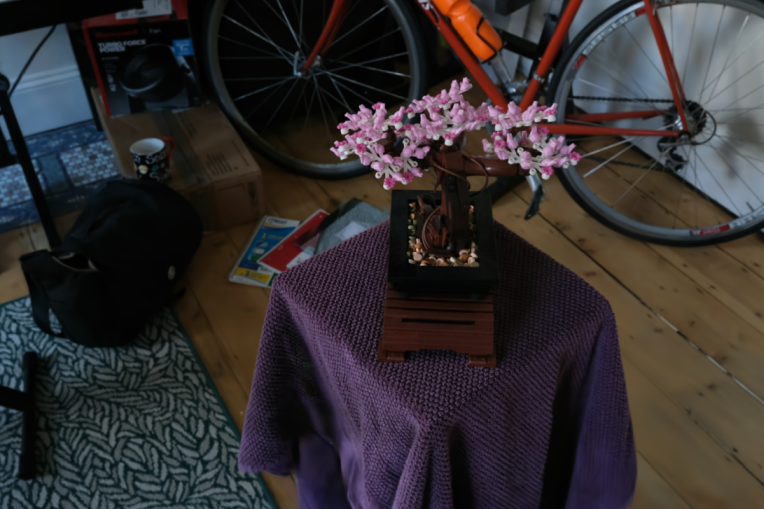}{2.9,1.75}{3,0.7}&
    \spyimg{0.19\textwidth}{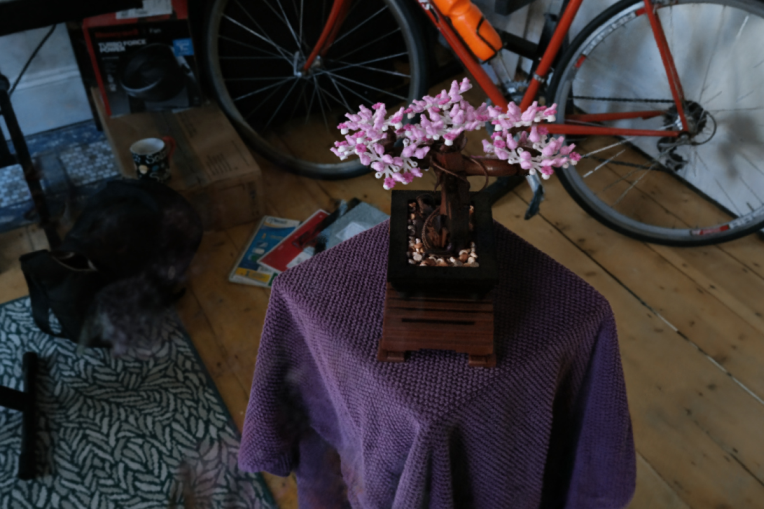}{2.9,1.75}{3,0.7}&
    \spyimg{0.19\textwidth}{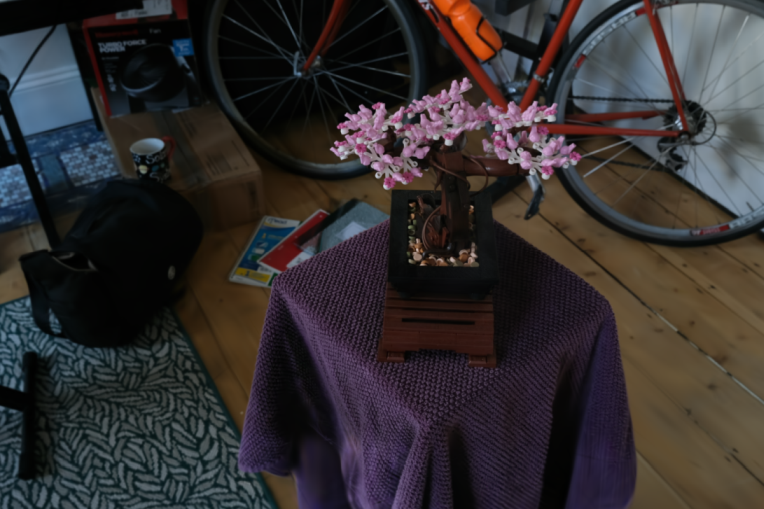}{2.9,1.75}{3,0.7}&
    \spyimg{0.19\textwidth}{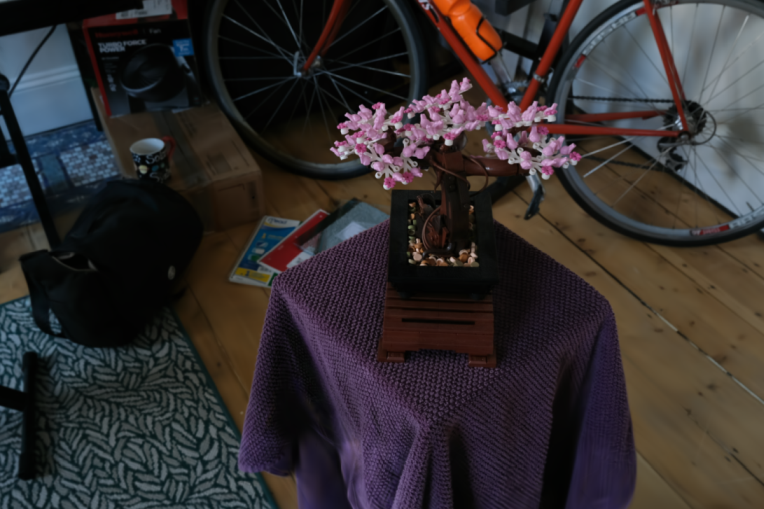}{2.9,1.75}{3,0.7}\\

    \spyimgred{0.19\textwidth}{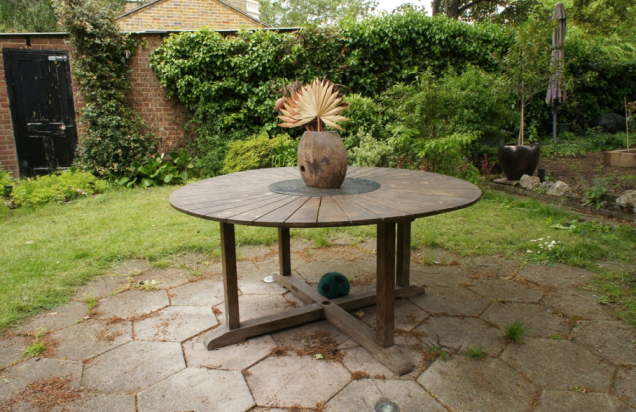}{0.65,1.85}{1.2,0.7}&
    \spyimgred{0.19\textwidth}{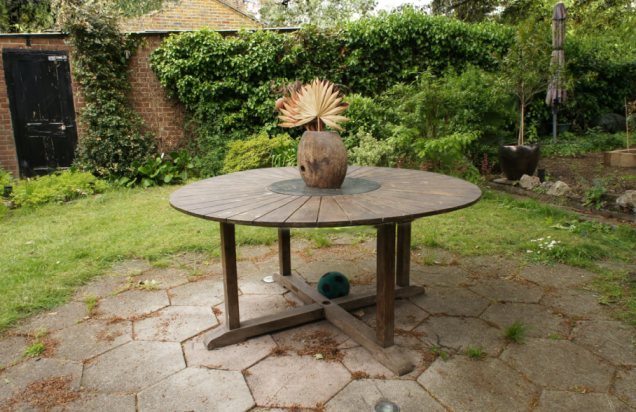}{0.65,1.85}{1.2,0.7}&
    \spyimgred{0.19\textwidth}{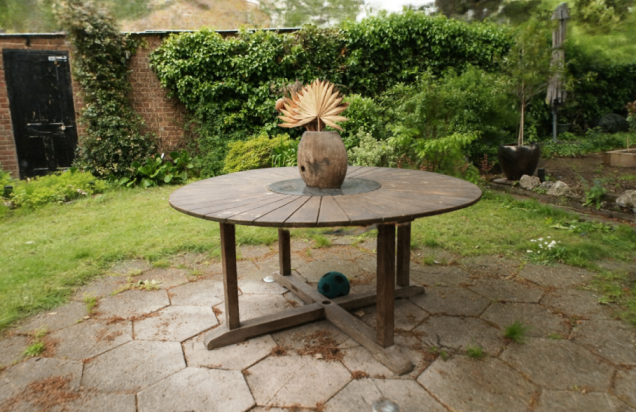}{0.65,1.85}{1.2,0.7}&
    \spyimgred{0.19\textwidth}{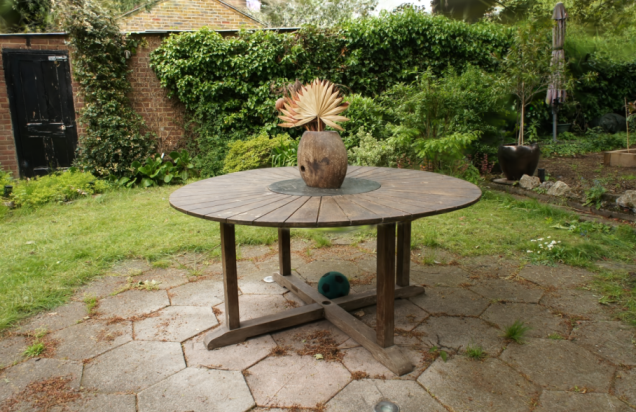}{0.65,1.85}{1.2,0.7}&
    \spyimgred{0.19\textwidth}{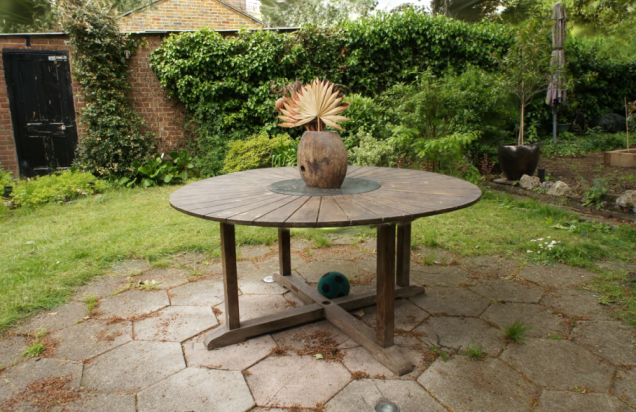}{0.65,1.85}{1.2,0.7}\\

    \spyimg{0.19\textwidth}{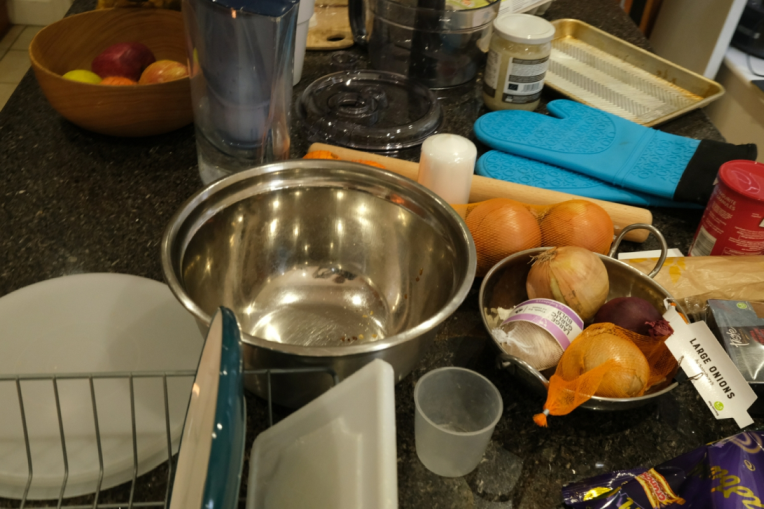}{0.5,0.5}{2.9,0.6}& 
    \spyimg{0.19\textwidth}{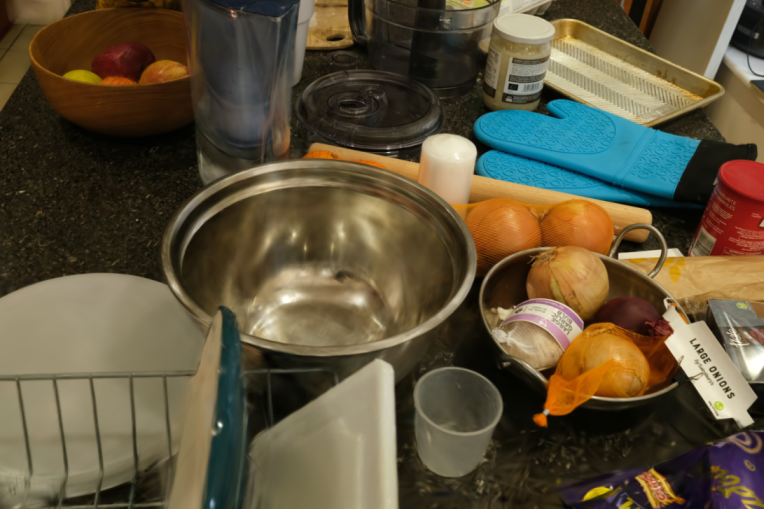}{0.5,0.5}{2.9,0.6}& 
    \spyimg{0.19\textwidth}{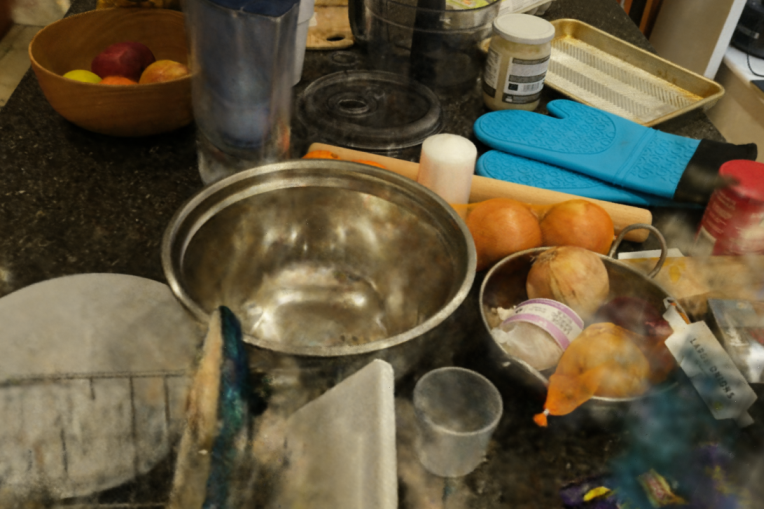}{0.5,0.5}{2.9,0.6}& 
    \spyimg{0.19\textwidth}{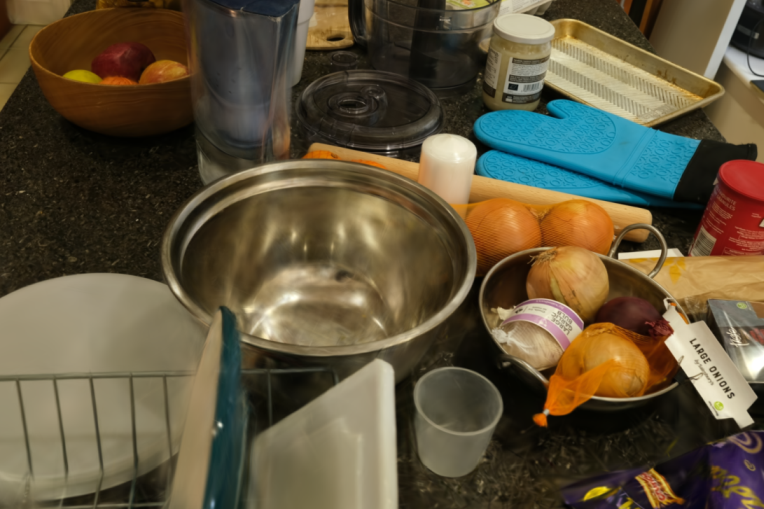}{0.5,0.5}{2.9,0.6}& 
    \spyimg{0.19\textwidth}{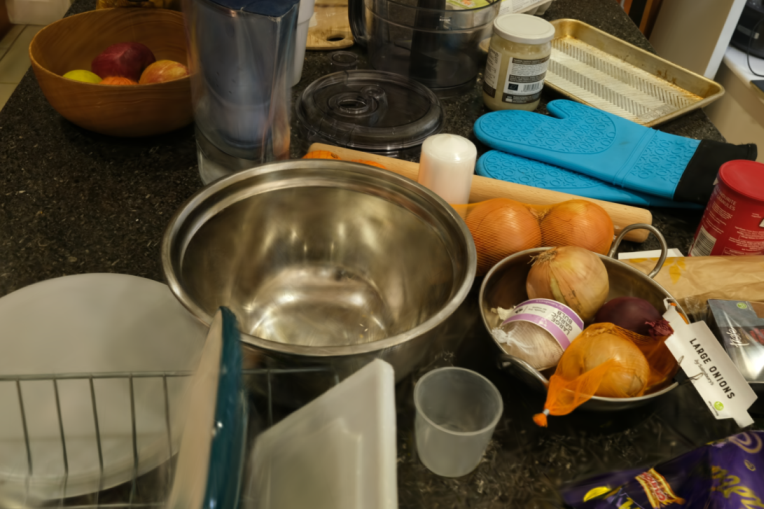}{0.5,0.5}{2.9,0.6}\\

    \spyimg{0.19\textwidth}{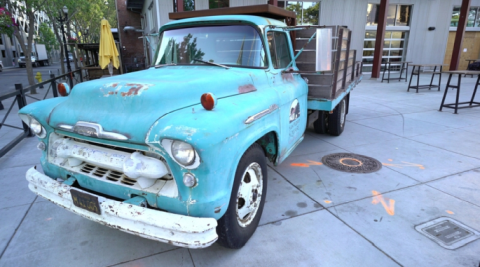}{0.2,1.0}{2.9,0.6}&
    \spyimg{0.19\textwidth}{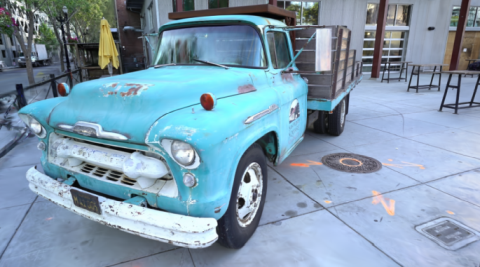}{0.2,1.0}{2.9,0.6}&
    \spyimg{0.19\textwidth}{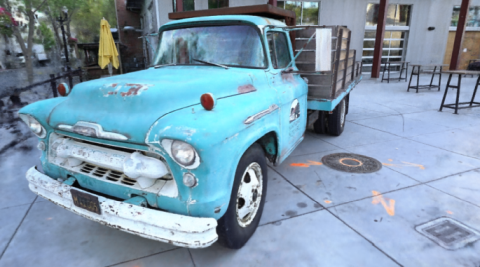}{0.2,1.0}{2.9,0.6}&
    \spyimg{0.19\textwidth}{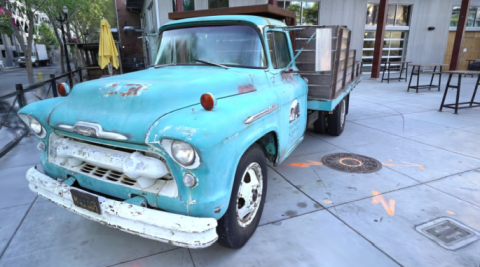}{0.2,1.0}{2.9,0.6}&
    \spyimg{0.19\textwidth}{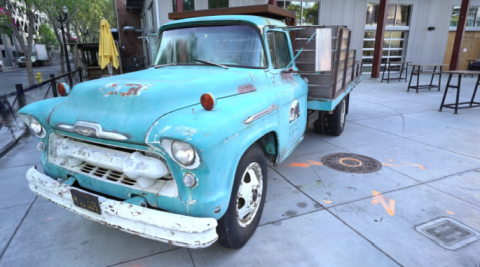}{0.2,1.0}{2.9,0.6}\\

    \spyimg{0.19\textwidth}{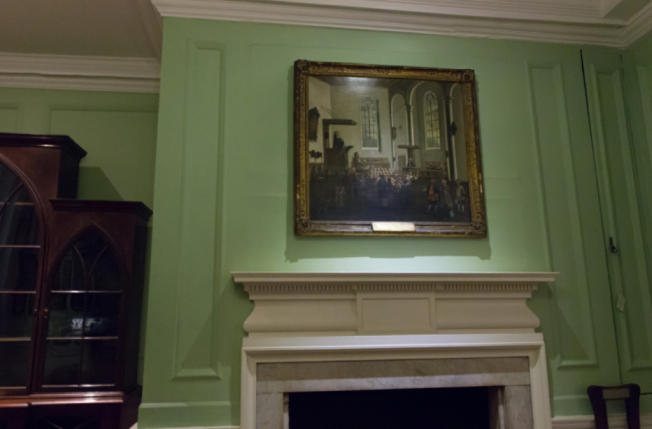}{2.9,1.85}{3,0.7}&
    \spyimg{0.19\textwidth}{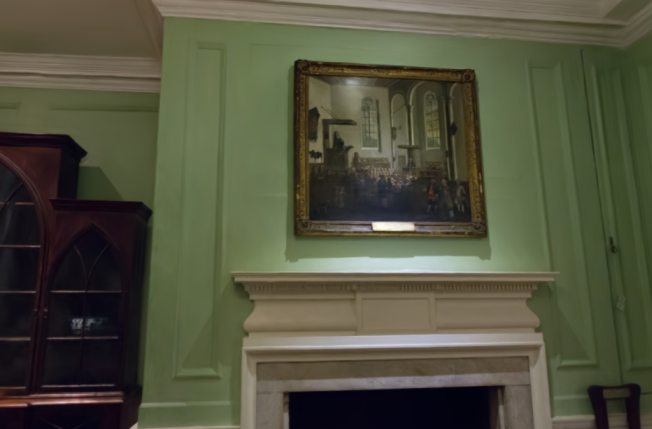}{2.9,1.85}{3,0.7}&
    \spyimg{0.19\textwidth}{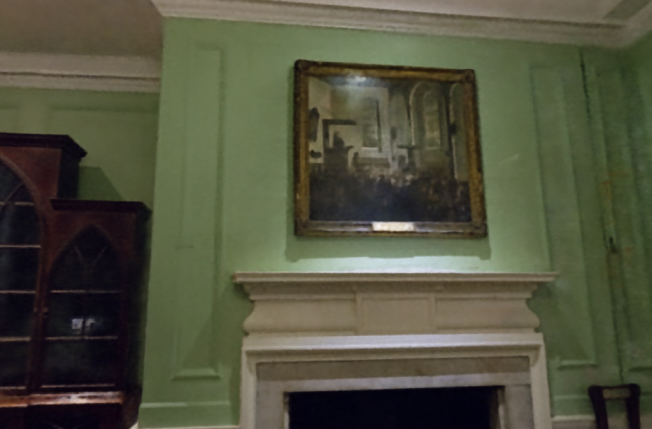}{2.9,1.85}{3,0.7}&
    \spyimg{0.19\textwidth}{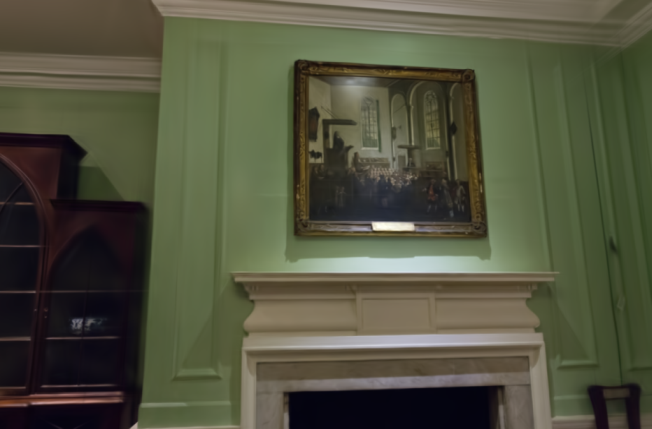}{2.9,1.85}{3,0.7}&
    \spyimg{0.19\textwidth}{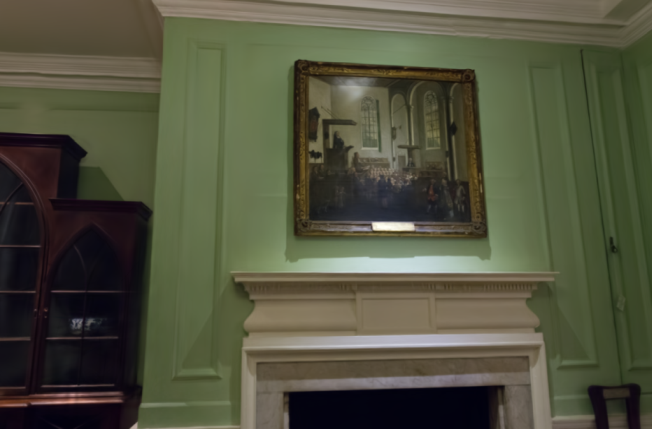}{2.9,1.85}{3,0.7}\\

    \spyimg{0.19\textwidth}{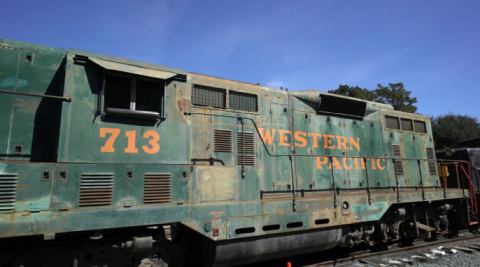}{1.4,1.1}{3,0.7} &
    \spyimg{0.19\textwidth}{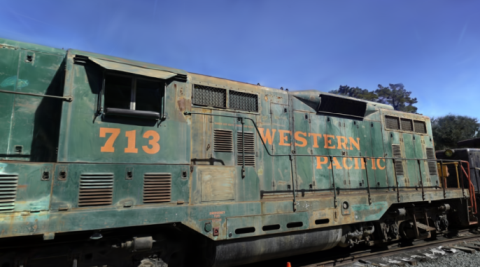}{1.4,1.1}{3,0.7} &
    \spyimg{0.19\textwidth}{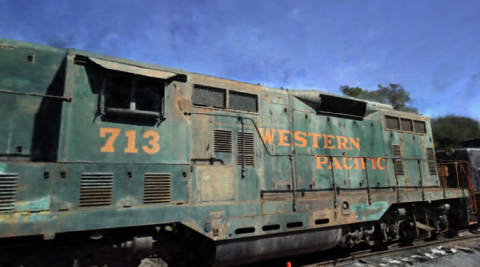}
    {1.4,1.1}{3,0.7} &
    \spyimg{0.19\textwidth}{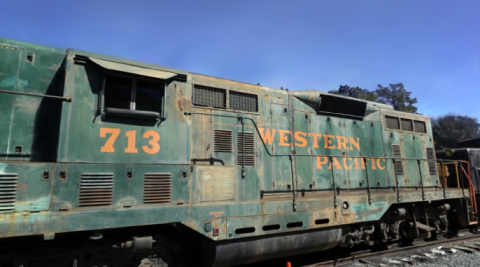}{1.4,1.1}{3,0.7} &
    \spyimg{0.19\textwidth}{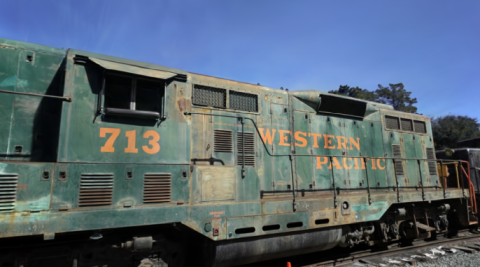}{1.4,1.1}{3,0.7} \\

\end{tabular}

    \caption{Qualitative comparison of results produced with our method in two budgeted scenarios to 3DGS, as well as Instant-NGP, whose training times match those of Ours. While the strictly budgeted scenario produces highly competitive results, a higher budget resolves occasional remaining blurry Gaussians.}
    \label{fig:fig1}
\end{figure*}

\subsection{Ablations}

Table~\ref{tab:ablations} examines the effect of individually removing several of our contributions on the scenes from Tanks\&Temples. This analysis is performed in the first budgeted scenario. 

\begin{wraptable}{R}[-0.5cm]{8cm}
\caption{Ablations on Tanks\&Temples.}
\label{tab:ablations}
\scalebox{0.74}{
\begin{tabular}{l|ccc|ccc|}
\multicolumn{1}{c}{}    & \multicolumn{3}{c|}{Truck} & \multicolumn{3}{c}{Train} \\ \cline{2-7} & \multicolumn{1}{c|}{PSNR}  & \multicolumn{1}{c|}{LPIPS} & Time & \multicolumn{1}{c|}{PSNR}  & \multicolumn{1}{c|}{LPIPS} & Time \\ \hline
Ours                 & \multicolumn{1}{c|}{\underline{25.20}} & \multicolumn{1}{c|}{\underline{0.165}} & \underline{7\,m}    & \multicolumn{1}{c|}{\underline{22.69}} & \multicolumn{1}{c|}{\underline{0.238}} & \underline{9\,m}   \\
--score-based sampling          & \multicolumn{1}{c|}{24.92} & \multicolumn{1}{c|}{0.189} & \textbf{6\,m}    & \multicolumn{1}{c|}{22.24} & \multicolumn{1}{c|}{0.246} & \textbf{8\,m} \\
--image loss          & \multicolumn{1}{c|}{24.94} & \multicolumn{1}{c|}{0.187} & 7\,m    & \multicolumn{1}{c|}{22.08} & \multicolumn{1}{c|}{0.242} & 9\,m   \\
--high opacity        & \multicolumn{1}{c|}{25.01} & \multicolumn{1}{c|}{0.174} & 7\,m    & \multicolumn{1}{c|}{22.29} & \multicolumn{1}{c|}{0.239} & 9\,m   \\
--reduce SH frequency & \multicolumn{1}{c|}{\textbf{25.39}} & \multicolumn{1}{c|}{\textbf{0.161}} & 9\,m   & \multicolumn{1}{c|}{\textbf{22.75}} & \multicolumn{1}{c|}{\textbf{0.235}} & 12\,m   \\
--per splat backward  & \multicolumn{1}{c|}{\underline{25.20}} & \multicolumn{1}{c|}{\underline{0.165}} & 14\,m   & \multicolumn{1}{c|}{\underline{22.69}} & \multicolumn{1}{c|}{\underline{0.238}} & 17\,m   \\
\hline
\end{tabular}
}
\end{wraptable} 

Note that all configurations yield the same number of Gaussians. However, omitting the consideration of image loss (or our score-based sampling altogether) from densification significantly harms quality. 

We observe a similar impact when omitting the use of high-opacity Gaussians. Reverting to the original SH update frequency can lead to minuscule quality improvements, but causes a fitting speed performance drop of up to 50\%. Replacing our per-splat backward pass with the original has an even higher performance cost, indicating the effectiveness of our optimizations. 
As an additional case study, Fig.~\ref{fig:teaser} ablates the quantitative effect on \textsc{Garden} when varying the available budget. We see a consistent improvement as budget increases, showing a clear correlation between provided budget and achieved image quality.
While our approach does not target the peculiarities of PyTorch, we note that our first budgeted scenario allows training with consistently less than 10\,GB VRAM---compact enough for a mid-range NVIDIA RTX 3080.

\begin{figure*}
    \centering
    \includegraphics[width=0.64\textwidth]{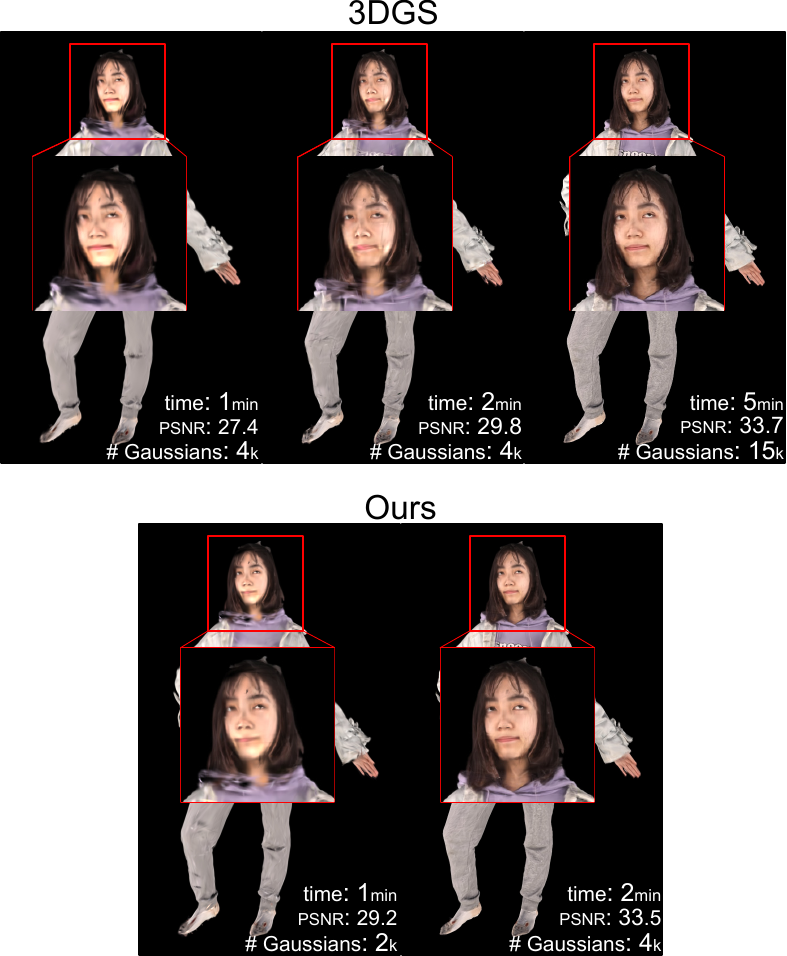}
	\caption{Demonstrating prioritization for guiding densification to regions of interest. We provide a region-of-interest mask for the face (detected with SegmentAnything \cite{kirillov2023segment}) on a model from the THuman dataset \cite{tao2021function4d} in the computation of $\textbf{S}_v$ and increase its weight to $10^3$. The above figure displays the quality of the facial region as measured via PSNR. We achieve equal scores much faster and with a fraction of the Gaussians used by 3DGS. This demonstrates the potential of our 3DGS budgeting for latency-constrained live scenarios: In a telepresence setting, we could prioritize the quality of the most frequently observed image regions---e.g., faces---and leave others under-sampled, without significantly degrading the user experience.}
	\label{fig:mask_experiment}
\end{figure*}

\section{Conclusion}
We have presented a highly efficient, splatting-based optimization technique for high-quality radiance fields. 
Our approach restrains the unpredictable behavior of the recent 3DGS technique, allowing for exact primitive budgeting, flexible sample steering, and highly improved resource efficiency, avoiding excessive peaks in training.
These properties generate new opportunities for optimizing novel-view synthesis in various environments, e.g., hardware-constrained and edge devices. Other potential applications include latency-constrained streaming services, where on-the-fly, interactive 3D reconstructions could be steered towards prioritizing salient regions of interest, such as faces (see Fig.~\ref{fig:mask_experiment}). 

Our contributions are complementary to ongoing 3DGS compression efforts, many of which could be applied to our reduced-size models to even greater effect.
While our approach is an important step towards low-cost, high-quality radiance fields, achieving optimal quality still requires a substantial sample count and meandering exploration as Gaussians move across the scene. We consider efficient search paths, occupancy predictions, and resolution of blind spots in scene reconstructions as exciting avenues for future work.

% \begin{figure*}

% %\scalebox{0.75}
% \centering
% \begin{minipage}{\textwidth}
% \centering
% {\Large 3DGS} 
% \\[0.3cm]
% \scalebox{0.79}{
% \begin{tikzpicture}[spy using outlines={red,magnification=4,size=4.1cm, connect spies}]
% \node[anchor=south west,inner sep=0] at (0,0) {\includegraphics[trim=0 34cm 0 5cm, clip, width=0.8\textwidth]{figures/body_experiment.pdf}};
% \spy on (2.2,7.1) in node [left] at (3.4, 4.25);
% \spy on (6.3,7.1) in node [left] at (8.22, 4.25);
% \spy on (10.5,7.1) in node [left] at (13, 4.25);
% \end{tikzpicture}
% }
% \end{minipage}

% % Adding vertical space to separate the tikzpictures

\bibliographystyle{unsrt}
\bibliography{sample-base}

\end{document}